%% file: main.tex
\documentclass{article}

\usepackage{microtype}
\usepackage{graphicx}
\usepackage{booktabs} 
\usepackage{xcolor,colortbl}

\usepackage{hyperref}

\usepackage[accepted]{icml2023}

\usepackage{amsmath}
\usepackage{amssymb}
\usepackage{mathtools}
\usepackage{amsthm}

\usepackage[capitalize,noabbrev]{cleveref}

\theoremstyle{plain}

\theoremstyle{definition}

\theoremstyle{remark}

\usepackage[textsize=tiny]{todonotes}

\icmltitlerunning{Spectral Maps for Learning on Subgraphs}

\input{math_commands.tex}

\usepackage{url}
\usepackage{placeins}
\usepackage{algorithm}
\usepackage{algorithmic}
\usepackage{array}
\usepackage{overpic}
\usepackage{wrapfig}
\usepackage{bm}
\usepackage{comment}
\usepackage{pgfplots}
\usepackage{pgf}
\usepackage{tikz}
\usepackage{soul}
\usepackage{tabstackengine}
\usepackage{mathtools}
\usepackage{multirow}
\usepackage{caption}
\usepackage{subcaption}
\usepackage{mdframed}
\usepackage{adjustbox}

\usepackage{tabularx}
\newcolumntype{L}[1]{>{\raggedright\arraybackslash}p{#1}}
\newcolumntype{C}[1]{>{\centering\arraybackslash}m{#1}}
\newcolumntype{a}{>{\centering\arraybackslash}X}
\newcolumntype{R}[1]{>{\raggedleft\arraybackslash}p{#1}}

\newcommand{\M}{\mathcal{M}}
\newcommand{\N}{\mathcal{N}}

\newcommand{\Gg}{G_{1}}
\newcommand{\Eg}{E_{1}}
\newcommand{\Vg}{V_{1}}

\newcommand{\Gs}{G_{2}}
\newcommand{\Es}{E_{2}}
\newcommand{\Vs}{V_{2}}
\newcommand{\Lap}{\mathcal{L}}

\newcommand{\EVECg}{\Phi_{1}}

\newcommand{\EVECs}{\Phi_{2}}

\newcommand{\EVEC}{\Phi}
\newcommand{\EVAL}{\Lambda}

\definecolor{mygray}{RGB}{120,120,120} 

\definecolor{revcolor}{RGB}{250,128,114} 

\definecolor{Gray}{gray}{0.85}

\newcommand{\first}[1]{\textcolor{red}{\textbf{#1}}}
\definecolor{red_second}{RGB}{255,99,71}
\newcommand{\second}[1]{\textcolor{red_second}{\textbf{#1}}}

\pgfplotsset{compat=1.17}
\begin{document}

\twocolumn[
\icmltitle{Spectral Maps for Learning on Subgraphs}



\icmlsetsymbol{equal}{*}

\begin{icmlauthorlist}
\icmlauthor{Marco Pegoraro}{Sap}
\icmlauthor{Riccardo Marin}{Tub}
\icmlauthor{Arianna Rampini}{Sap}
\icmlauthor{Simone Melzi}{Mil}
\icmlauthor{Luca Cosmo}{Ca}
\icmlauthor{Emanuele Rodolà}{Sap}
\end{icmlauthorlist}

\icmlaffiliation{Sap}{Sapienza University of Rome, Italy }
\icmlaffiliation{Mil}{University of Milano-Bicocca, Italy}
\icmlaffiliation{Ca}{Ca’ Foscari University of Venice, Italy}
\icmlaffiliation{Tub}{University of T{\"u}bingen, Germany}

\icmlcorrespondingauthor{}{}

\icmlkeywords{Spectral theory, Learning on graphs, Subgraphs, Maps Representation}
\vskip 0.3in
]

\begin{abstract}
In graph learning, maps between graphs and their subgraphs frequently arise. For instance, when coarsening or rewiring operations are present along the pipeline, one needs to keep track of the corresponding nodes between the original and modified graphs. Classically, these maps are represented as binary node-to-node correspondence matrices, and used as-is to transfer node-wise features between the graphs. In this paper, we argue that simply changing this map representation can bring notable benefits to graph learning tasks. Drawing inspiration from recent progress in geometry processing, we introduce a spectral representation for maps that is easy to integrate into existing graph learning models. This spectral representation is a compact and straightforward plug-in replacement, and is robust to topological changes of the graphs. Remarkably, the representation exhibits structural properties that make it interpretable, drawing an analogy with recent results on smooth manifolds. We demonstrate the benefits of incorporating spectral maps in graph learning pipelines, addressing scenarios where a node-to-node map is not well defined, or in the absence of exact isomorphism. Our approach bears practical benefits in knowledge distillation and hierarchical learning, where we show comparable or improved performance at a fraction of the computational cost.

\end{abstract}


\input{./sections/introduction}

\input{./sections/related}
\input{./sections/background}

\input{sections/sigtransf.tex}

\input{sections/applications.tex}
\input{sections/analysis.tex}
\input{./sections/conclusion}
\section*{Acknowledgements}
This work is supported by the ERC Starting Grant No. 802554 (SPECGEO), the SAPIENZA BE-FOR-ERC 2020 Grant (NONLINFMAPS), NVIDIA Academic Hardware Grant Program, and an Alexander von Humboldt Foundation Research Fellowship.

\bibliography{eg_bib}
\bibliographystyle{icml2023}

\newpage
\appendix
\onecolumn
\input{sections/Appendix/interpretation}
\input{sections/Appendix/exp_details}
\input{sections/Appendix/GZoom_eigs.tex}


\end{document}

%% file: math_commands.tex
\def\1{\bm{1}}

\DeclareMathAlphabet{\mathsfit}{\encodingdefault}{\sfdefault}{m}{sl}
\SetMathAlphabet{\mathsfit}{bold}{\encodingdefault}{\sfdefault}{bx}{n}



%% file: sections/introduction.tex
\section{Introduction}
\label{sec:introduction}

Graph learning offers a powerful set of techniques for understanding complex data, which often call for downsampling or rewiring operations to improve on scalability and performance. One common approach is to perform computations and training on a partial or modified version of the graph, rather than the entire graph. For example, computationally expensive operations can be performed on a coarsened version of the graph, as demonstrated in works such as \cite{deng2020graphzoom}. Additionally, graph rewiring, which directly modifies the connectivity, creates an even more challenging scenario \cite{chan2016rewire,PE_rewiring}. In these settings, a crucial aspect that is often taken for granted is the data transfer between graphs and their subgraphs. Recent studies have shown that transferring information such as positional encoding from a graph to its rewired versions can improve GNN performance \cite{PE_rewiring}, highlighting the importance of effectively transferring information between graphs. However, this task remains challenging in many scenarios, particularly when the involved graphs are not isomorphic. Although correspondences between nodes are often provided, utilizing these correspondences as they are may not always be the optimal solution, leaving room for further improvement.

In this paper, we propose to shift to a {\em spectral} representation as a way to compactly encode maps between graphs and subgraphs in graph learning pipelines. The new representation is a straightforward replacement into existing models; it is easy to compute, has a regularizing behavior leading to improved downstream performance, and bears a natural structure that is easy to interpret. From a technical standpoint, the map representation is obtained via a change of basis with respect to the eigenvectors of the graph Laplacian. This idea, introduced a decade ago in the field of geometry processing under then name of {\em functional maps}~\citep{ovsjanikov2012functional}, has led to notable advancements in several tasks in graphics and vision. However, the potential application of this concept in graph learning has not been explored so far.

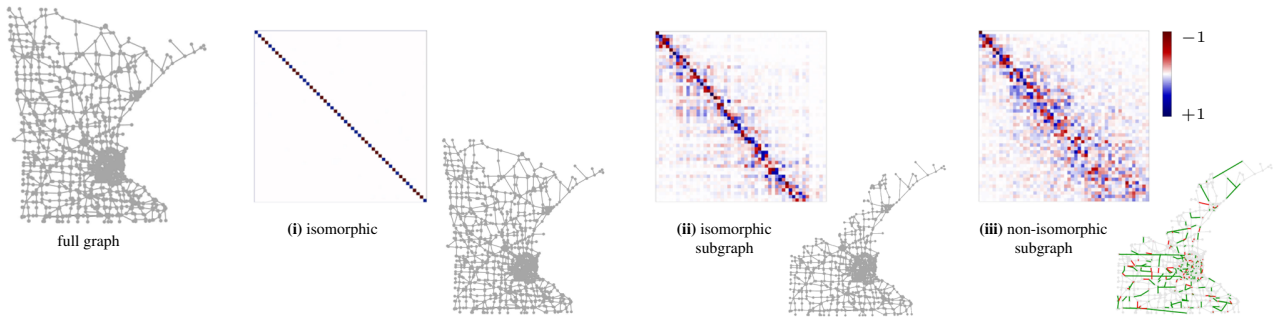
\begin{figure*}[t!]
\centering
\begin{overpic}
[trim=0cm 0cm 0cm 0cm,clip,width=0.99\linewidth, grid=false]{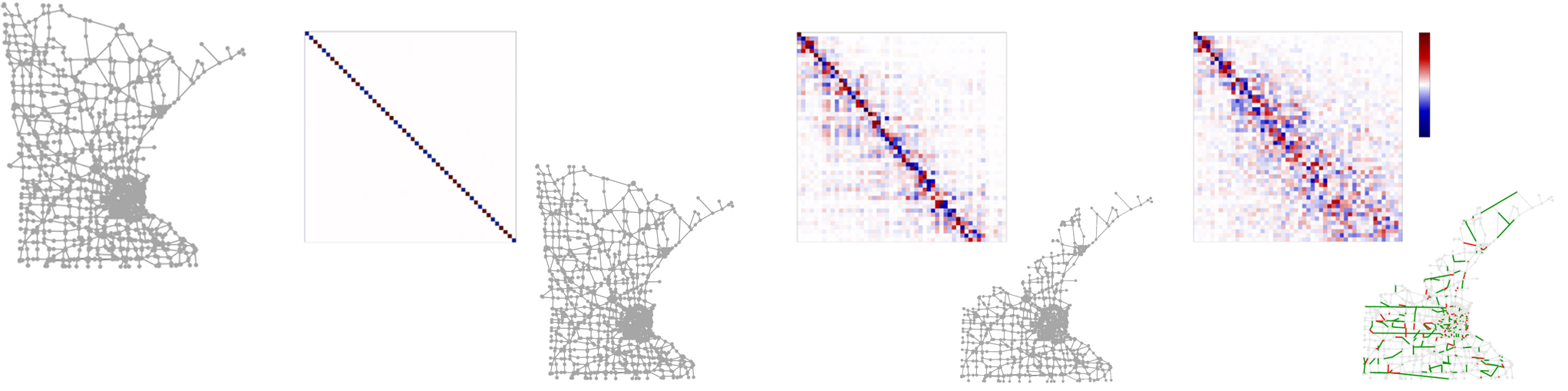}
\put(4,6){\tiny full graph}
\put(22,7){\tiny \textbf{(i)} isomorphic}
\put(52.5,6){\parbox{.5in}{\centering \tiny \textbf{(ii)} isomorphic subgraph }}
\put(76,6){\parbox{.7in}{\centering \tiny \textbf{(iii)} non-isomorphic subgraph }}
\put(92,16){\tiny $+1$}
\put(92,22){\tiny $-1$}
\end{overpic}
\caption{\label{fig:minnesota_detail}Spectral maps 
between a full graph (depicted on the left) and three different graphs, respectively: an isomorphic graph (\textbf{i}), an isomorphic subgraph containing $80\%$ of the original nodes (\textbf{ii}), and a {\em non}-isomorphic subgraph obtained by randomly rewiring the former (\textbf{iii}). The green edges are novel and randomly added ($10\%$ of the total), the red edges are randomly removed ($10\%$ of the total). The map representation still maintains a visible structure despite the significant changes of connectivity that span several hops.}
\end{figure*}



We summarize our main contributions as follows:
\begin{itemize}
\item We propose the adoption of spectral representations for maps between graphs and {\em subgraphs}. For the first time, we show that such maps exhibit a special structure in their coefficients, capturing the similarity between the Laplacian eigenspaces of the two graphs. 
\item We conduct an empirical examination of the structure of the functional map across a diverse range of graphs and in various scenarios of partiality, including sub-isomorphic and non-isomorphic graphs. Our findings demonstrate that the map exhibits a distinct structure in these contexts; see Figure~\ref{fig:minnesota_detail} for examples.
\item We focus on the problem of feature transfer and include experiments showing practical applications, {such as hierarchical embedding and knowledge distillation,} on graphs spanning a few dozen to tens of thousands of nodes. In terms of performance and computational complexity, we also demonstrate key benefits.
\end{itemize}


The present article is structured as follows: We provide an overview of existing literature on graph-to-graph mappings and their significance in learning and non-learning procedures in Section 2. Section 3 introduces the spectral representation and its mathematical formalism, which are fundamental for defining the spectral maps. Section 4 illustrates the utility of these maps by showcasing two examples of learning pipelines where their application results in superior performance. The validity of the properties inherited by these representations is evaluated in various scenarios in Section 5. Finally, in Section 6, we summarize the key findings and discuss potential avenues for future research.


%% file: sections/related.tex
\section{Related work}
\label{sec:related}
In this section, we review the literature on the use of maps in graph learning models, where our method has a direct relevance.


\textbf{Maps for graph learning.}
Transferring information between non-isomorphic graphs is a challenging problem in graph learning. 
This is especially relevant in scenarios such as domain adaptation \cite{pilanci2020domain}, meta-learning \cite{yang2022data}, and federated learning \cite{zhang2021subgraph}, where the information collected on a set of graphs needs to be transferred to other graphs. 
In this paper we focus on the problem of {\em representing} maps between graphs, given a (possibly partial) node-to-node correspondence; however, it is worth noting that there are several methods that tackle the complementary problem of determining a correspondence~\cite{IsoRank,GRASP,PALE} when the latter is not provided as input.


\textbf{Hierarchical graph embedding.}
Many learning-based graph embedding algorithms, such as DeepWalk \cite{DeepWalk} and node2vec \cite{node2vec}, do not scale to large graphs and struggle to capture long-distance global relationships \cite{chen2018harp}. To overcome these problems, recent works \cite{chen2018harp, deng2020graphzoom, liang2021mile} proposed to compute a hierarchy of coarsened graphs on which to compute the embeddings, and then lift the values up to the original graph. 
In this framework, an important step is the propagation of embeddings through the coarsened graphs, which requires a proper refinement step to ensure the quality of the final embedding. In particular, ensuring smooth propagation between levels has been identified as a crucial element in enhancing performance. In our experiments, we show how the spectral representation can be easily adapted for this step, with beneficial effects on the graph embedding task; we refer to Section \ref{sec:hiercEmb} for a detailed evaluation.

\textbf{Knowledge distillation on graphs.}
The goal of knowledge distillation is to transfer information from a large model to a smaller one~\cite{hinton2015distilling}. Recently, this framework has been extended to graphs \cite{yang2020distilling,chen2020self,yang2021extract}. Specifically, \cite{gkd} introduced the concept of geometric knowledge distillation, which involves transferring graph topology information extracted by a GNN model from a graph $G$ (Teacher) to a target GNN model; importantly, the target GNN only has access to a partial view of $G$ (Student). In this paper, we address this task by adopting the spectral representation to enforce the similarity between the intermediate representation learned by the teacher and the student (Section \ref{sec:GKD}).

%% file: sections/background.tex
\section{\label{sec:background} Background on spectral representation}

\begin{figure}
    \centering
    \begin{overpic}
    [trim=0cm 0.1cm 0cm 1cm,clip,width=0.999\linewidth, grid=false]{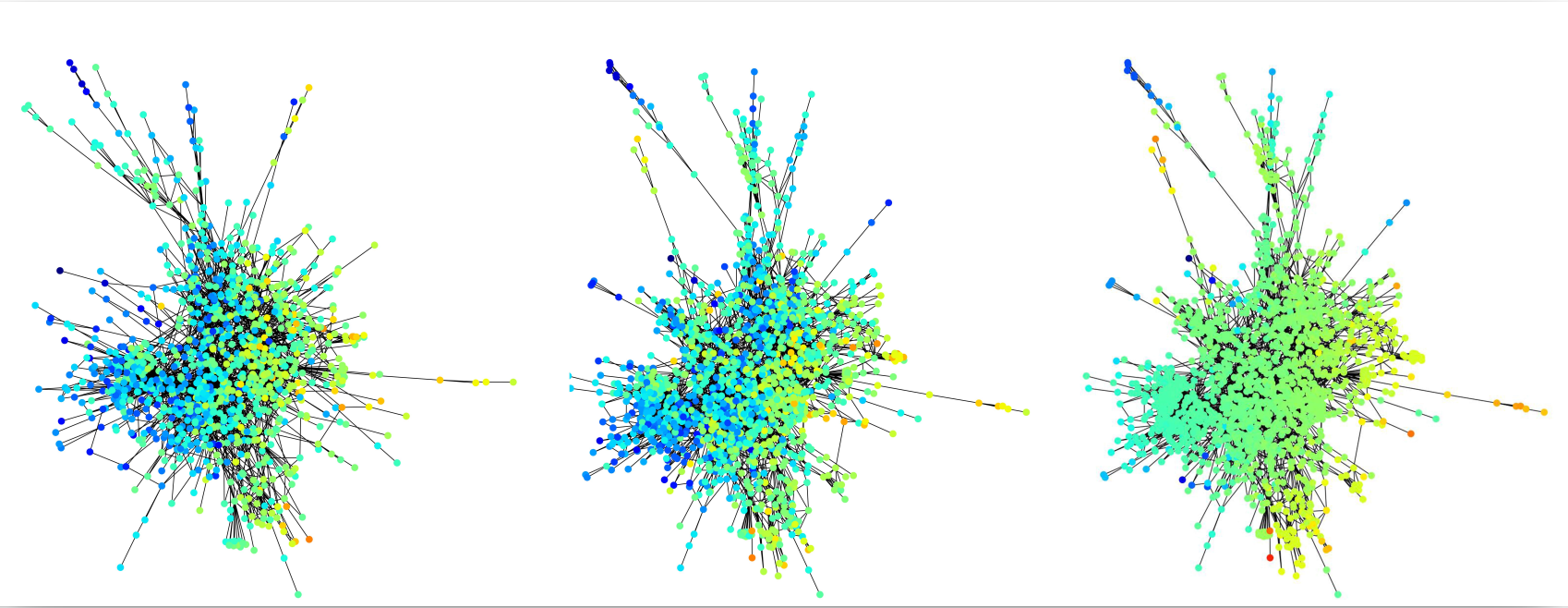}
    \put(6,-2){\small Node2Vec}
    \put(42,-2){\small GraphZoom}
    \put(80,-2){\small Ours}
    \end{overpic}
    \caption{\label{fig:GZoom sigtransf}
    The Node2Vec embedding is, from left to right, applied to the coarsened graph, transferred to the full graph with GraphZoom and with the spectral map. We remark the smoothing effect of the spectral representation.
    }
\end{figure}

\paragraph{Graphs and Laplacian eigenvectors.}
We consider undirected, unweighted graphs $G = (V,E)$ with nodes $V$ 
and edges $E \subseteq V \times V$. 
We denote as $A \in \{ 0, 1\}^{\vert V \vert \times \vert V \vert}$ the adjacency matrix of $G$, which is a binary matrix where $A(i,j) = 1$ if an edge connects node $i$ to node $j$, and $A(i,j) = 0$ otherwise.

The {symmetric normalized Laplacian} for $G$ is defined as the square matrix $\Lap = I - D^{-\frac{1}{2}}A D^{-\frac{1}{2}}$, 
where $D$ is a diagonal matrix of node degrees, with entries $D(i,i) = \sum_{j = 1}^{\vert V \vert} A(i,j)$.
%
%
This linear operator is symmetric and positive semi-definite; it admits an eigendecomposition $\Lap = \EVEC \EVAL \EVEC^{\top}$, where $\EVAL$ is a diagonal matrix that contains the eigenvalues, and $\EVEC$ is a matrix having as columns the corresponding eigenvectors concatenated side by side.
Throughout this paper, we assume the eigenvalues (and corresponding eigenvectors) to be sorted in non-descending order $0=\lambda_1\leq \lambda_2 \leq \ldots \leq 2$; this is important for interpreting the spectral maps that we define in the sequel.

Each eigenvector $\phi_{l}$ for $l = 1, \ldots , \vert V \vert$ has length $\vert V \vert$, and can be interpreted as a scalar function defined on the nodes of the graph; for this reason, we will refer to them as eigen{\em functions}.
The eigenfunctions form an orthonormal basis for the space of functions defined on the graph nodes (i.e. $\EVEC^{\top} \EVEC = Id$).
One may consider a subset of eigenfunctions, namely those associated with the $k$ 
 smallest eigenvalues, to compactly approximate a graph signal. 

\textbf{Functional maps for graphs.}
The representation we propose directly derives from the functional maps framework for smooth manifolds~\cite{ovsjanikov2012functional}, extended to the graph setting in~\cite{wang2019functional,GRASP}. 

Consider two graphs $\Gg = (\Vg,\Eg)$ and $\Gs = (\Vs,\Es)$ and a binary matrix $S$ encoding a node-to-node map $T:\Gs\to\Gg$. Applying an orthogonal change of basis w.r.t. bases $\Phi_1,\Phi_2$, we get to the representation:
\begin{align}\label{eq:Cg}
C = \Phi_2^\top S \Phi_1\,,
\end{align}
where $\Phi_1 \in \mathbb{R}^{\vert \Vg \vert \times k}, \Phi_2 \in \mathbb{R}^{\vert \Vs \vert \times k}$ contain the first $k$ eigenvectors of the symmetrically normalized graph Laplacians of $G_1$ and $G_2$ respectively, and $S \in \mathbb{R}^{\vert \Vg \vert \times \vert \Vs \vert }$ is a matrix encoding the node-to-node correspondence.
This matrix $C$ is easy to compute by simple matrix multiplication.
The size of $C$ does {\em not} depend on the number of points in $\Gg$ and $\Gs$, but only on the number $k$ of basis functions.
In other words, $C$ represents the linear transformation that maps the coefficients of any given function $f:\Vg\to\mathbb{R}$ expressed as linear combination of $\Phi_1$, to coefficients of a corresponding function $g:\Vs\to\mathbb{R}$ expressed in the eigenbasis $\Phi_2$. 


Graph nodes may often come with numerical attributes encoding user identities in social networks, or positional encodings. We can model such data as a collection of functions $f: \Vg \rightarrow \mathbb{R}$. 
From Equation \ref{eq:Cg} we can transfer a  function $f$ from $\Gg$ to $\Gs$ applying the following formula:
%
\begin{align}
\hat{g} = \EVECs C \EVECg^\top f \,,
\label{eq:transfer}
\end{align}
{where $\EVECg^\top$ projects $f$ in its coefficients, $C$ apply the spectral transfer, $\EVECs$ reconstucts the transfered signal $\hat{g}$.}


%% file: sections/sigtransf.tex

%% file: sections/applications.tex
\section{\label{sec:application} Applications on subgraphs}

From now on we consider the setting where we are given a graph $\Gg$ and a possibly noisy subgraph $\Gs = (\Vs,\Es)$ of $\Gg$, such that $\Vs \subseteq \Vg$ and $\Es \subseteq \Eg$. 
%
In this case, Equation~\ref{eq:Cg} still holds.
Note that in some cases, one may decide to invert the roles of the graphs $\Gg$ and $\Gs$, as in Section \ref{sec:hiercEmb}. This does not affect the spectral representation of the map.

\subsection{\label{sec:motiv}Motivation}
Our motivation starts from the observation that in many practical cases, the eigenspaces of the normalized graph Laplacian are well preserved under {\em non-isomorphic} transformations of the graph, including strong partiality, topological perturbations, and edge rewiring.

 According to Equation~\eqref{eq:Cg}, each coefficient $c_{ij}$ of $C$ corresponds to a dot product between $\phi^2_i$ and $S \phi^1_j$; this measures the correlation {\em at corresponding nodes} between a Laplacian eigenvector $\phi^2_i$ of $G_2$, and a Laplacian eigenvector $\phi^1_j$ of $G_1$. Each eigenvector $\phi^1_j$ is expressed as a linear combination of images through $S$ of the eigenvectors $\phi^2_i$ (i.e. $S \phi^2_i$), and the combination coefficients are stored in column $j$ of $C$.

To explain with an example how the structure of $C$ relates to the graph eigenspaces, consider the example of the Minnesota graph in Figure~\ref{fig:minnesota_detail}.
Suppose we map the full graph to its permuted version (\textbf{i}). In this case, the two graphs have the same eigenspaces due to the permutation equivariance of Laplacian eigenvectors.
Thus, the matrix $C$ is diagonal with $\pm 1$ along the diagonal because $c_{ij}=0$ for $i\neq j$ (due to orthogonality of the eigenvectors), while $c_{ii}=\pm 1$ (due to the sign ambiguity of the eigenvectors). In the case of repeated eigenvalues, one may observe small blocks of coefficients along the diagonal due to the non-uniqueness of the choice of the eigenvectors spanning high-dimensional eigenspaces.
When we map the full graph to its subgraph (\textbf{ii}), the two graphs have partially similar eigenspaces, meaning that the inner products between $\phi^2_i$ and $S\phi^1_j$ tend to be close to zero and close to $\pm 1$, but not exactly equal. The matrix $C$ has a {\em sparse} structure but is not necessarily diagonal. This is because the eigenvectors on the subgraph correlate with those of the full graph at different indices $i\neq j$ -- unlike the full-to-full case, where the correspondence happens at $i=j$. 
Therefore the spectral map matrices are not necessarily diagonal but may present a different sparsity structure which depends on the particular graph and subgraph.
\setlength{\columnsep}{15pt}
\setlength{\intextsep}{3pt}
\begin{wrapfigure}[8]{r}{0.45\linewidth}
\vspace{-0.15cm}
\begin{center}
\begin{overpic}
[trim=0cm 0cm 0cm 0cm,clip,width=0.95\linewidth, grid=false]{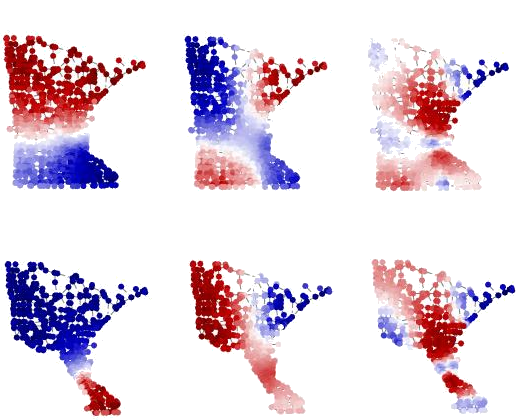}
\put(-8,60){\tiny $\Gg$}
\put(9,76){\tiny $\mathbf{2}$}
\put(45,76){\tiny $\mathbf{5}$}
\put(80,76){\tiny $\mathbf{10}$}
\put(-8,15){\tiny $\Gs$}
\put(9,32){\tiny $\mathbf{2}$}
\put(45,32){\tiny $\mathbf{4}$}
\put(80,32){\tiny $\mathbf{8}$}
\end{overpic}
\end{center}
\vspace{1.8cm}
\end{wrapfigure}
The inset figure shows an example of this phenomenon. The eigenfunctions of the complete graph $\Gg$, and those of the subgraph $\Gs$ still correlate even if not necessarily at the same index {(see pair 5-4) and the correlation may not be exact (see pair 10-8);  the extent to which the eigenfunctions correlate is captured precisely by the structure of $C$.
In particular, we can see that the values of the Laplacian eigenfunctions stay approximately the same (up to sign, in the case of simple spectrum) at the nodes that are not directly involved in the perturbation -- which is to say that the eigenvectors of the partial graph $G_2$, encoded in $\EVECs$, correlate strongly with the those of $G_1$, encoded in $\EVECg$. 
 To the best of our knowledge, this observation is not trivial and has not been reported before. 
This simple fact leads to the following important observation that is central to our contribution:

\textit{The spectral representation allows us to represent the same (or similar) subspace of smooth functions by truncating the functional representation at the first k eigenfunctions}

Since eigenfunctions align well, we can exploit the spectral maps and the properties they inherit on the representation and transfer of signals. While classically, maps are represented as binary matrices $S$ whose dimensions scale quadratically with the number of nodes in the graphs, this observation allows us to use the spectral map as a compact and sparse representation that still provides an efficient way of transferring information between graphs. Furthermore, as we will show in the rest of this section, the properties inherited from this representation provide advantages in applications. 
In all the following experiments, we inject the spectral representation in learning procedures only where information transfer is needed, leaving the rest of the pipeline unchanged. In Section \ref{sec:analysis}, we show extensive empirical evidence of this behavior and describe its practical consequences. 

\input{sections/Tabels/TAB_GZoom.tex}
\subsection{\label{sec:hiercEmb}Hierarchical Graph Embedding}

Hierarchical Graph Embedding aims to learn a graph embedding considering a hierarchy of coarsened graphs. First, each level of the hierarchy is constructed from the original graph. Then, an embedding is computed on the last level (i.e. smallest subgraph), and finally it is lifted up to the original graph. In this case, the correspondence between the original graph and its subgraphs is given by construction. 

In this section, we show that transferring the embeddings across the hierarchy levels via the spectral map is beneficial in the applications. To this end, we consider the state-of-the-art Hierarchical Graph embedding approach GraphZoom \cite{deng2020graphzoom} to compute the coarsened graphs.
Then, we transfer the embedding using Equation \eqref{eq:transfer} in the reverse direction,
as we transfer the signal from the subgraph to the full graph. Equation \eqref{eq:transfer} still holds, but $\Gg$ is now the subgraph, and $\Gs$ is the full graph. As the results show, this swap does not affect the performance of the map.
The spectral map is computed from the ground truth correspondences obtained during the coarsening phase.

To evaluate performance, we tackle the task of node classification.
The classification is performed by a linear logit regression model that takes as input the  embedding lifted up through the hierarchy of graphs. 
As done in \cite{deng2020graphzoom}, we consider two levels of coarsening. 
Table \ref{tab:GZoom} shows the node classification accuracy of our method compared to GraphZoom \cite{deng2020graphzoom} and a baseline $n2n$.
We consider here three graphs (Cora, Citeseer and Pubmed) and three embedding algorithms (node2vec \cite{node2vec}, DeepWalk \cite{DeepWalk} and GraphSAGE \cite{graphSAGE}), similarly to \cite{deng2020graphzoom}.
The only hyperparameter of our approach is the number of eigenvectors $k$ employed in the spectral map computation. 
In the last two rows of Table \ref{tab:GZoom}, we report the best accuracy obtained at varying percentage of eigenvectors (\textit{Ours}) and the accuracy obtained using the fixed percentage $5\%$ (\textit{Ours fixed}).
 The GraphZoom method is replicated using the parameters provided by the official code repository.


\textbf{Regularizing behavior.}
Using $k\ll n$ eigenvectors in the construction of $C$ has a regularizing effect on the map, akin to a low-pass filtering of the correspondence. 
Figure \ref{fig:GZoom sigtransf} shows on Cora an example of embedding transferred from the coarsest level to the next in the hierarchy. We use a spectral map computed with the first $5\%$ eigenvectors (last column). The spectral transfer performs an evident smoothing on the embedding, compared to Graphzoom (middle column).
As further evidence, we plot the classification accuracy at varying percentages of eigenvectors (Figure \ref{fig:HGEeigs} in Appendix \ref{app:GZoomeigs}). Importantly, 
the performance peaks with few eigenvectors and then decreases when increasing the number of eigenvectors up to the complete base.
In particular, when we use {\em all} the eigenvectors $\Phi_1$ and $\Phi_2$ to construct $C$, Equation~\ref{eq:Cg} corresponds to an orthogonal change of basis; therefore, the representations $S$ and $C$ are equivalent and have the same dimensions.
Truncating the bases to the first $k_1$ and $k_2$ eigenvectors, as described in Appendix~\ref{sec:nume_eig}, yields a low-rank approximation $C \approx S$.
%
In signal processing terms, we see the matrix $C$ as a {\em band-limited} representation of the node-to-node correspondence $S$. 

It was already demonstrated in \cite{nt2019revisiting} and \cite{li2019label} that smoothing signals can improve performance on graphs. And our results validate this idea once again.
Even if, the regularizing effect is desirable in many cases but is traded off for a loss in accuracy if a precise node-to-node correspondence is desired. On the one hand, if the map $C$ is used to transfer a smooth signal (e.g. node-wise features like spectral positional encodings or carrying semantic information depending on the data), then the loss in accuracy is negligible since Laplacian eigenvectors are optimal for representing smooth signals~\citep{aflalo2015optimality}; on the other hand, transferring non-smooth signals via a small $C$ has the effect of filtering out the high frequencies. If high frequencies are desired, it is often sufficient just to increase the values of $k_1, k_2$, leading to a bigger matrix $C$.
In Section \ref{sec:sigtransf}, we analyse how the transferred signal changes at different numbers of eigenvectors.

\input{sections/Tabels/TAB_GKD.tex}
\subsection{\label{sec:GKD}Geometric Knowledge Distillation}
The aim of Geometric Knowledge Distillation \cite{gkd} is to transfer topological knowledge from a teacher model to a student model, which has only a partial vision of the graph. In particular, the teacher model is trained on $\Gg=(\Vg,\Eg)$ and the student on $\Gs=(\Vs,\Es)$.
In this scenario, the spectral map can be used to align the features that the teacher and student models are learning. For this purpose, we define the following loss:
\begin{align}
|| C \EVECg^T x_t - \EVECs^T x_s ||
\label{eq:fmaploss}
\end{align}
where $x_t \in \mathrm{R}^{|\Vg| \times d}$ and $x_s \in \mathrm{R}^{|\Vs| \times d}$ are the features computed by the teacher and the student, $\EVECg \in \mathrm{R}^{|\Vg| \times k}$ and $\EVECs \in \mathrm{R}^{|\Vs| \times k}$ are the eigenvectors on the teacher and student graph respectively and $C \in \mathrm{R}^{k \times k}$ is the spectral map between the two graphs. We remark that both $\EVECg$ and $\EVECs$ are precomputed before training time. 

In Table \ref{tab:GKD}, we compare the student trained with Equation \eqref{eq:fmaploss} with the methods proposed in \cite{gkd}: gaussian kernel (GKD-G), random kernel (GKD-R), sigmoid kernel (GKD-S) and parametric kernel (PGKD). 
In the first three rows, we also report the performance of the ORACLE model (trained and tested on $\Gg$), TEACHER (trained on $\Gg$ and tested on $\Gs$) and STUDENT (trained and tested on $\Gs$).  
We consider two settings: \textit{node-aware} where the subgraph's nodes are a subset of the full graph's node $\Vs \subset \Vg$; \textit{edge-aware} where the subgraph's edges are a subset of the full graph's edges, but the nodes are the same $\Es \subset \Eg$ and $\Vs = \Vg$. In both cases, the partiality considered is $50\%$.
We report the node classification accuracy over three runs with random initialization. We train all the models for 500 epochs.

In all the datasets the spectral representation reaches an accuracy comparable with at least one of the methods proposed by \cite{gkd}. In particular in the node-aware setting, the spectral map always perform as the first or second best method. 
The edge aware setting corresponde to a non-isomorphic transofrmation of the graph. In this case, as we will show in Section \ref{sec:RobRewire}, the spectral representation still holds a compact representation but at higher percentages of partiality the correlation between the eigenspaces of the graphs tends to be weaker. As the results show, this can damage the performance of the spectral representation even if not drastically. We believe this is can lead to further investigation on the non-isomorphic mapping of graphs.

The spectral representation is also a faster method than \cite{gkd}. Since the eigenvector can be precomputed, at training time the only additional expense is a simple matrix multiplication. Overall, the spectral representation can reach a speedup of $200\%$ compared to \cite{gkd}. In Appendix \ref{app:GKDadd} we show the full table with the computation time per epoch and speed ups.

%% file: sections/Tabels/TAB_GZoom.tex
\begin{table}[t]
\caption{\label{tab:GZoom} \textit{Hierarchical embedding}: Mean classification accuracy on the task of node classification. 
}
\centering
\footnotesize
\setlength{\tabcolsep}{3pt}
\begin{tabular}{c| c c c}
            & Graphzoom &  Ours (\% eigs) & Ours (fixed) \\ \toprule
            \multicolumn{4}{c}{Node2Vec} \\ \toprule
   Cora     & 0.77 &   \first{0.79} (10\%) &   0.78 (5\%)  \\
   Citeseer & 0.64 &  \first{0.67} (2.5\%) & \first{0.67} (5\%) \\
   Pubmed   & 0.79 &   \first{0.80} (10\%) & 0.79 (5\%) \\ \toprule
        \multicolumn{4}{c}{Graph Walk} \\ \toprule
   Cora     &  0.76    & \first{0.79} (15\%) & 0.77 (5\%) \\
   Citeseer & 0.65  & \first{0.68}(2.5\%) & 0.67 (5\%) \\
   Pubmed   & 0.78  & \first{0.80} (5\%)&   \first{0.80} (5\%) \\ \toprule
        \multicolumn{4}{c}{GraphSAGE} \\ \toprule
   Cora     &  0.72  &  \first{0.74} (3\%) & 0.68 (5\%) \\
   Citeseer & 0.55     & \first{0.59} (60\%) & 0.56 (5\%) \\
   Pubmed   &  \first{0.74}    &  \first{0.74} (5\%) & \first{0.74} (5\%)  \\ 
\end{tabular}
\end{table}

%% file: sections/Tabels/TAB_GKD.tex
\begin{table*}[t]

\caption{\label{tab:GKD} \textit{Knowledge distillation}: Results of node classification accuracy over multiple runs. We compare the spectra representation (Ours) with the methods proposed by \cite{gkd} (GKD-G, GKD-R, GKD-S, PGKD). For Ours we also report the percentages of eigenvectors used in the spectral map.}
\begin{subtable}{1\textwidth}
\centering
\tiny
\caption{\label{tab:GKD_node} Node-aware knowledge setting}
\begin{tabular}{l|c|c|c|c|c|c}
              & Cora & Citeseer & {Amazon-photo} & Amazon-computer & Coauthor-cs & Pubmed  \\ \toprule
\rowcolor{Gray}
ORACLE& $87.74 \pm 1.41 $& $70.35 \pm 1.30 $& ${93.00} \pm 0.63 $& ${90.90} \pm 0.56 $& ${92.89} \pm 0.41 $& $86.33 \pm 0.25 $\\
TEACHER& $85.52 \pm 0.51 $& $68.91 \pm 1.62 $& $91.93 \pm 0.11 $& $89.34 \pm 0.42 $& $92.25 \pm 0.45 $& $85.16 \pm 0.56 $\\
STUDENT& $82.87 \pm 2.09 $& $69.19 \pm 2.27 $& $92.30 \pm 0.40 $& $85.41 \pm 2.30 $& $85.67 \pm 2.28 $& $85.09 \pm 0.16 $\\ \midrule
GKD-G & $\second{88.08} \pm 0.95 $& $\second{71.07} \pm 0.36 $& $92.02 \pm 0.41 $& $\second{90.13} \pm 0.19 $& $\first{92.75} \pm 0.61 $& $85.97 \pm 0.15 $\\
GKD-R& $\first{88.13} \pm 1.32 $& $70.83 \pm 1.89 $& $\second{92.44} \pm 0.43 $& $88.25 \pm 0.79 $& $\second{92.13} \pm 0.86 $& $86.05 \pm 0.42 $\\
GKD-S& $87.64 \pm 0.56 $& $\first{71.15} \pm 0.88 $& $\second{92.44} \pm 0.67 $& $89.95 \pm 0.50 $& $92.26 \pm 0.37 $& $86.01 \pm 0.27 $\\
PGKD& $86.71 \pm 0.82 $& $68.95 \pm 0.55 $& $92.21 \pm 0.67 $& $89.80 \pm 0.12 $& $92.02 \pm 0.14 $& $\second{86.36} \pm 0.34 $\\ \midrule
Ours (\% eigs) & $\second{88.08} \pm 1.11 (12\%)$& $\first{71.15} \pm 1.13 (50\%)$& $\first{92.84} \pm 0.28 (25\%)$& $\first{90.94} \pm 0.50 (50\%)$& $\second{92.13} \pm 0.37 (50\%)$& $\first{86.42} \pm 0.39 (50\%)$\\
\end{tabular}
\end{subtable}
\begin{subtable}{1\textwidth}
\centering
\tiny
\caption{\label{tab:GKD_edge} Edge-aware knowledge setting}
\begin{tabular}{l|c|c|c|c|c|c}
              & Cora & Citeseer & Amazon-photo & Amazon-computer & Coauthor-cs & Pubmed  \\ \hline
\rowcolor{Gray}
ORACLE& ${87.74 }\pm 1.41 $& $70.35 \pm 1.30 $& ${93.00} \pm 0.63 $& ${90.90} \pm 0.56 $& ${92.89} \pm 0.41 $& ${86.33} \pm 0.25 $\\
TEACHER& $81.88 \pm 1.49 $& $67.11 \pm 2.05 $& $90.57 \pm 1.04 $& $87.47 \pm 0.26 $& $91.14 \pm 0.54 $& $83.19 \pm 0.49 $\\
STUDENT& $82.82 \pm 0.31 $& $70.47 \pm 1.42 $& $\first{92.42} \pm 0.54 $& $77.86 \pm 3.20 $& $83.38 \pm 1.48 $& $84.52 \pm 0.31 $\\ \midrule
GKD-G& $\first{87.54} \pm 0.23 $& $\first{71.51} \pm 0.82 $& $91.76 \pm 0.60 $& $\second{89.53} \pm 0.12 $& $\first{91.98} \pm 0.07 $& $\second{86.09} \pm 0.14 $\\
GKD-R& $86.76 \pm 1.48 $& $71.11 \pm 0.70 $& $\second{92.12 }\pm 0.36 $& $88.29 \pm 0.58 $& $91.73 \pm 0.35 $& $86.06 \pm 0.64 $\\
GKD-S& $\second{87.05} \pm 1.20 $& $71.31 \pm 2.65 $& $92.00 \pm 0.53 $& $88.49 \pm 0.64 $& $91.51 \pm 0.47 $& $86.07 \pm 0.43 $\\
PGKD& $86.21 \pm 0.56 $& $69.59 \pm 0.68 $& $\first{92.42} \pm 0.31 $& $89.30 \pm 0.61 $& $91.65 \pm 0.25 $& $\first{86.86} \pm 0.48 $\\ \midrule
Ours (\% eigs) & $86.26 \pm 0.39 (4\%)$& $\second{71.47} \pm 0.62 (4\%)$& $92.14 \pm 0.24 (25\%)$& $\first{90.16} \pm 0.46 (50\%)$& $\second{91.80} \pm 0.33 (50\%)$& $85.81 \pm 0.10 (50\%)$\\
\end{tabular}
\end{subtable}
\end{table*}

%% file: sections/analysis.tex
\section{\label{sec:analysis} Empirical results and analysis}
So far we have seen how the spectral representation can be easily plugged in into existing pipelines showing competitive performances.
In this section, we analyse the structure of the spectral map under different kind of partialities to give further insights on its benefits.

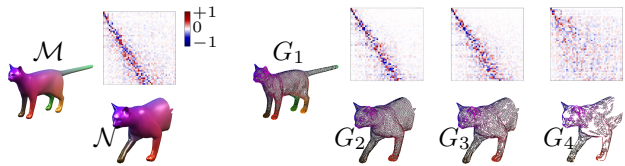
\begin{figure}[t!]
\vspace{0.2cm}
\centering
\begin{overpic}
[trim=0cm 0cm 0cm 0cm,clip,width=0.99\linewidth]{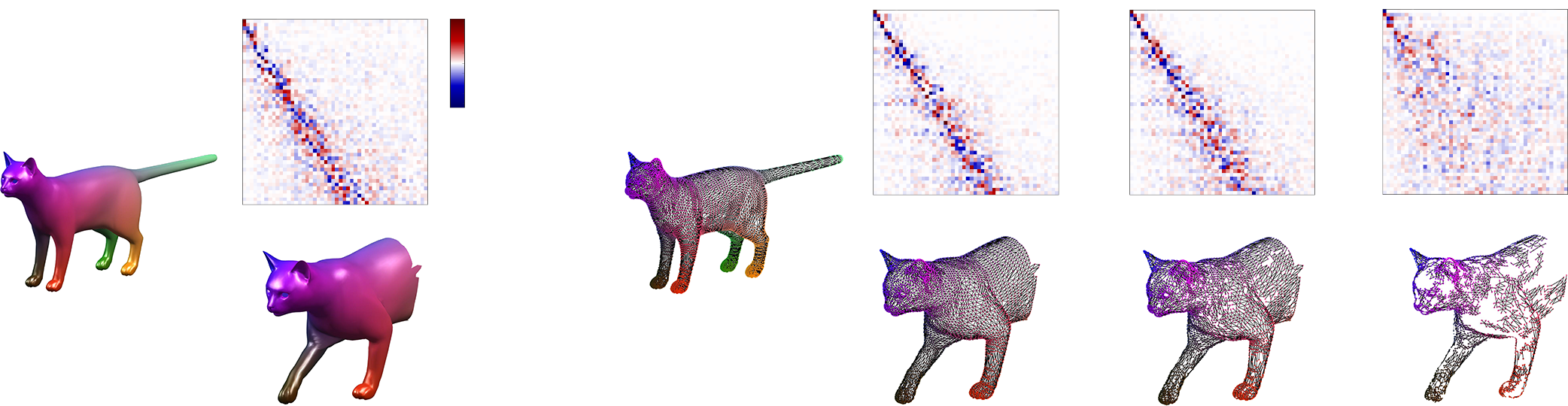}
\put(4,17){$\mathcal{M}$}
\put(14,3){$\mathcal{N}$}
\put(43,17){$G_1$}
\put(53,3){$G_2$}
\put(70,3){$G_3$}
\put(87,3){$G_4$}
\put(30,24){\tiny $+1$}
\put(30,21.5){\tiny $0$}
\put(30,19){\tiny $-1$}
\vspace{0.1cm}
\end{overpic}
\caption{\label{fig:cats}A spectral map between a full and a partial surface (left) compared to the spectral maps between a graph and three different subgraphs (right).}
\end{figure}

\input{sections/Figs/robust_rewire}
\subsection{Map structure}
{In Section \ref{sec:motiv} we highlighted how the most direct consequence of this preservation of eigenspaces is reflected in the structure of the spectral map $C$.}
In 3D geometry processing, a similar behavior was observed for the discrete Laplace-Beltrami operator under partiality transformations
~\citep{rodola2017partial,postolache20}; however, their theoretical analysis assumes the data to be Riemannian surfaces with a smooth metric -- an assumption that does {\em not} hold in the case of general graphs. 
%
We refer to Appendix~\ref{sec:smooth} for further details.

{In Figure~\ref{fig:cats} we show several examples of matrix $C$ for different subgraphs.} 
On the left we show the functional map matrix between a smooth surface $\mathcal{M}$ and a deformed part $\mathcal{N}$: the slanted-diagonal structure suggests that the eigenspaces of $\mathcal{M}$ are mostly preserved in $\mathcal{N}$. On the right, we show the spectral map matrices between a graph $G_1$ and different subgraphs: $G_2$ is obtained by removing $40\%$ of the nodes of $G_1$, while $G_3,G_4$ are obtained by removing $55\%$ and $80\%$ of the edges from $G_2$ respectively. The slanted-diagonal structure can still be observed and gets dispersed only at very high partiality. In the graphs, corresponding nodes have the same color.
The slanted-diagonal structure of the map between $\mathcal{M}$ and $\mathcal{N}$ is explained by an application of Weyl's law to 2-dimensional Riemannian manifolds, see \citealp[Eq. 9]{rodola2017partial} and Appendix \ref{sec:smooth}. However, there is no theoretical counterpart to explain the map structure between $G_1$ and its subgraphs, due to the complete absence of metric information about the underlying surface: the eigenfunctions are computed {\em purely} from the graph connectivity. Yet, the diagonal structure is preserved even under rather dense removal of edges, suggesting deeper algebraic implications.
%



One might legitimately ask whether the presence of a structure in the maps of Figure~\ref{fig:cats} is due to the specific choice of the data, where the subgraphs derive from a 3D mesh (although the normalized graph Laplacian dismisses any edge length information) and where the type of partiality resembles a neat `cut' (although we also perform random edge removals).
However, the same behavior is also observed with abstract graphs, as we show with CORA~\citep{cora} in Figure~\ref{fig:graphsMaps}, and with the datasets PPI0~\citep{PPI}, Amazon Photo~\citep{AmazonCoBuy} and Amazon Computer~\citep{AmazonCoBuy} in Figure~\ref{fig:amazon} of the Appendix.

\input{sections/Figs/minnecora}
\input{sections/Tabels/TAB_transfer}
\subsection{\label{sec:RobRewire}Non-isomorphic subgraphs}
%
In many practical settings, there are cases where the subgraph $G_2$ is contained in the bigger graph $G_1$ only up to some topological alterations; for example, 
{in the graph learning literature, topological perturbations frequently occur due to noise in the data, or are explicitly obtained by rewiring operations~\citep{flow21} or adversarial attacks~\citep{jin21} among others.}

{In Figure~\ref{fig:minnesota_detail}, we show the spectral map between Minnesota and a subgraph after rewiring (\textbf{iii}). We still observe a correspondence between the eigenvectors of the full graph and those of the subgraph. 
The spectral map has a sparse pattern, but it loosens up as the topological modifications increase.
For this to be true, we expect that small changes in graph connectivity lead to small changes in the matrix coefficients. See Appendix \ref{app:rob_rewire} for the formal definition.
}


{In Figure~\ref{fig:robustRewiring}, we evaluate the changes of the spectral map at increasing percentages of rewiring of a subgraph.
We consider six graphs and compute a subgraph from each one. Then, we apply small incremental changes to the topology of the subgraphs, with increments of $3\%$ of the total number of edges; the changes are performed by removing and adding random edges, obtaining new subgraphs $G_i$.
The plot on the right shows how much the spectral map representation is affected by the increasing topological changes compared to adding Gaussian noise. In all the cases, the rewiring produces less variation in the spectral map than in adding Gaussian noise. In particular, the functional representation is more robust on larger graphs, such as cat (10000 nodes) or citeseer (2120 nodes), while on smaller graphs such as QM9 (29 nodes) and Karate (34 nodes), removing or adding an edge has a more significant impact. 
This observation demonstrates the effectiveness of the spectral representation, especially on larger graphs.
In Appendix \ref{app:rob_rewire}, we show the complete qualitative analysis; while in the Supplementary Materials, we push this experiment to stronger rewiring.
}

{All the remarks so far directly depend on graph connectivity, and it is hard to find analogies for smooth surfaces.}
We conjecture that local topological transformations of a graph, while they can certainly induce strong transformations of {\em some} of its Laplacian eigenspaces (similar to single-point perturbations on planar manifolds, see~\cite{filoche2012universal}), are less likely to distort {\em all} the eigenspaces at once. This way, the spectral map matrix tends to maintain its global structure intact and exhibits local perturbations.




\subsection{\label{sec:sigtransf}Signal Transfer}
In Section \ref{sec:hiercEmb} and \ref{sec:GKD}, we leveraged Equation \ref{eq:transfer} to transfer information between graphs.
To better understand how the spectral representation afflicts the transferred signal, in Figure~\ref{fig:func_transfer}, we analyze the spectral map transfer performance while increasing the number of eigenfunctions used for the map representation.
We evaluate the fidelity of the transferred signal with the Root Mean Squared Error between the transferred signal $\hat{g}$ and the ground truth signal $g$ (obtained via the ground truth node-to-node correspondence):
\begin{equation}
    RMSE = \sqrt{\frac{1}{n}{\sum^n_{i}{(g(i)-\hat{g}(i))^2}}} \,,
\end{equation}
where $n$ is the number of nodes in the subgraph.
We consider pairs composed of the original graph and a series of subgraphs extracted according to a semantic criterion, e.g., nodes belonging to the same class or nodes connected by the same edge type. Motivated by the results from \cite{PE_rewiring}, we transfer the Random Walk Positional Encoding \citep{rwPE} computed on the full graphs to the subgraphs. We normalize each dimension of the node features of the original graph to exhibit zero mean and unitary standard deviation throughout all the nodes and then transfer this signal through Equation \ref{eq:transfer}.
In Figure~\ref{fig:func_transfer}, we can see how the Root Mean Squared Error between the spectral map and the ground truth transfer decreases as the number of eigenfunctions increases. 
In particular, the error is almost steady between $30\%$ and 75\%. 
This demonstrates the convenience of using fewer eigenvectors. 
The qualitative examples on the left of Figure \ref{fig:func_transfer} portray the transferred signal on PPI0. The transfer reaches a good approximation at 1\% of the eigenfunctions, while at $10\%$ and 75\% they are almost identical. This behaviour demonstrates that using a compact representation with few eigenvectors can approximate the signal well. 
In Appendix \ref{sec:nume_eig}, we show more experiments with different number of eigenfunctions.

%% file: sections/Figs/robust_rewire.tex
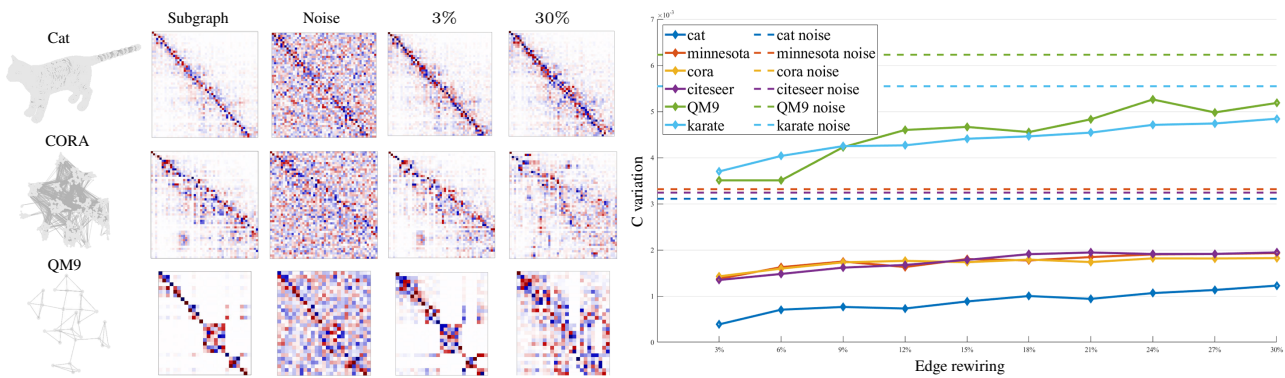
\begin{figure*}[t!]
\centering
\begin{overpic}
[trim=0cm 0cm 0cm 0cm,clip,width=0.999\linewidth, grid=false]{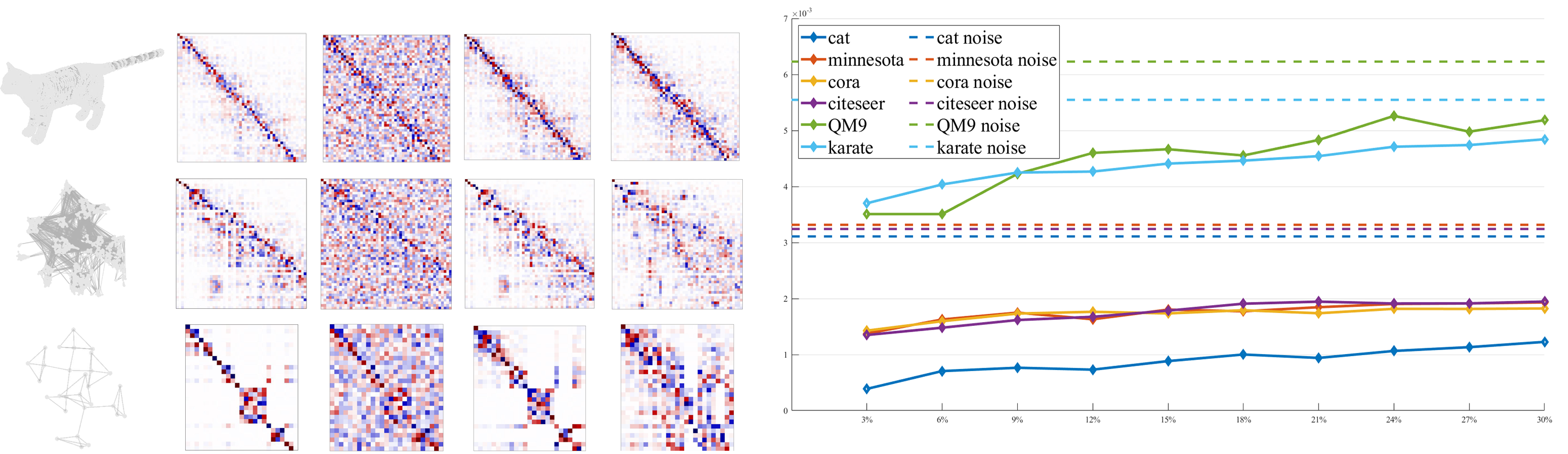}
\put(3.2,26){\tiny Cat}
\put(3,18){\tiny CORA}
\put(3.2,8.3){\tiny QM9}
\put(70.5,0.5){\tiny Edge rewiring}
\put(48.5,14){\tiny \rotatebox[origin=c]{90}{C variation}}
\put(12.5,27.5){\tiny Subgraph}
\put(23,27.5){\tiny Noise}
\put(33,27.5){\tiny $3\%$}
\put(41,27.5){\tiny $30\%$}
\end{overpic}
\caption{\label{fig:robustRewiring} Robustness of the map to the simultaneous action of partiality and rewiring of the subgraph. We compare the addition of gaussian noise ($\mu=0$;$\sigma=0.2$) with the impact of increasing rewiring (from 3\% to 30\% of the total number of edges) on the functional map $C$ of size $50\times50$. On the left, we plot three graphs with their functional map: no rewiring (Subgraph), the addition of gaussian noise (Noise), 3\% of edges rewired (3\%), and 30\% of edges rewired (30\%). On the right, we plot the variation of $C$ at different percentages of rewiring (solid lines) and with the addition of noise (dashed lines) for each graph.}
\end{figure*}

%% file: sections/Figs/minnecora.tex
\begin{figure}[t]
\centering
\begin{overpic}
[trim=0cm 0.5cm 0cm 0cm,clip,width=1\linewidth,grid=false]{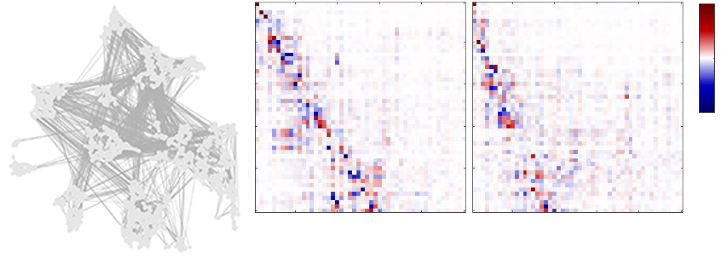}
%
\put(45.5,3.0){\tiny {\em CORA}}
\put(41.2,0.5){\tiny {class removal}}
\put(75.1,3.0){\tiny {\em CORA}}
\put(72.3,0.5){\tiny {random cut}}
\end{overpic}

\caption{\label{fig:graphsMaps}Spectral maps between CORA and two different subgraphs.}
\end{figure}

%% file: sections/Tabels/TAB_transfer.tex
\begin{figure*}[h]
     \centering
\begin{overpic}
[clip,width=0.57\linewidth,grid=false, trim=0 0 0 -1cm]{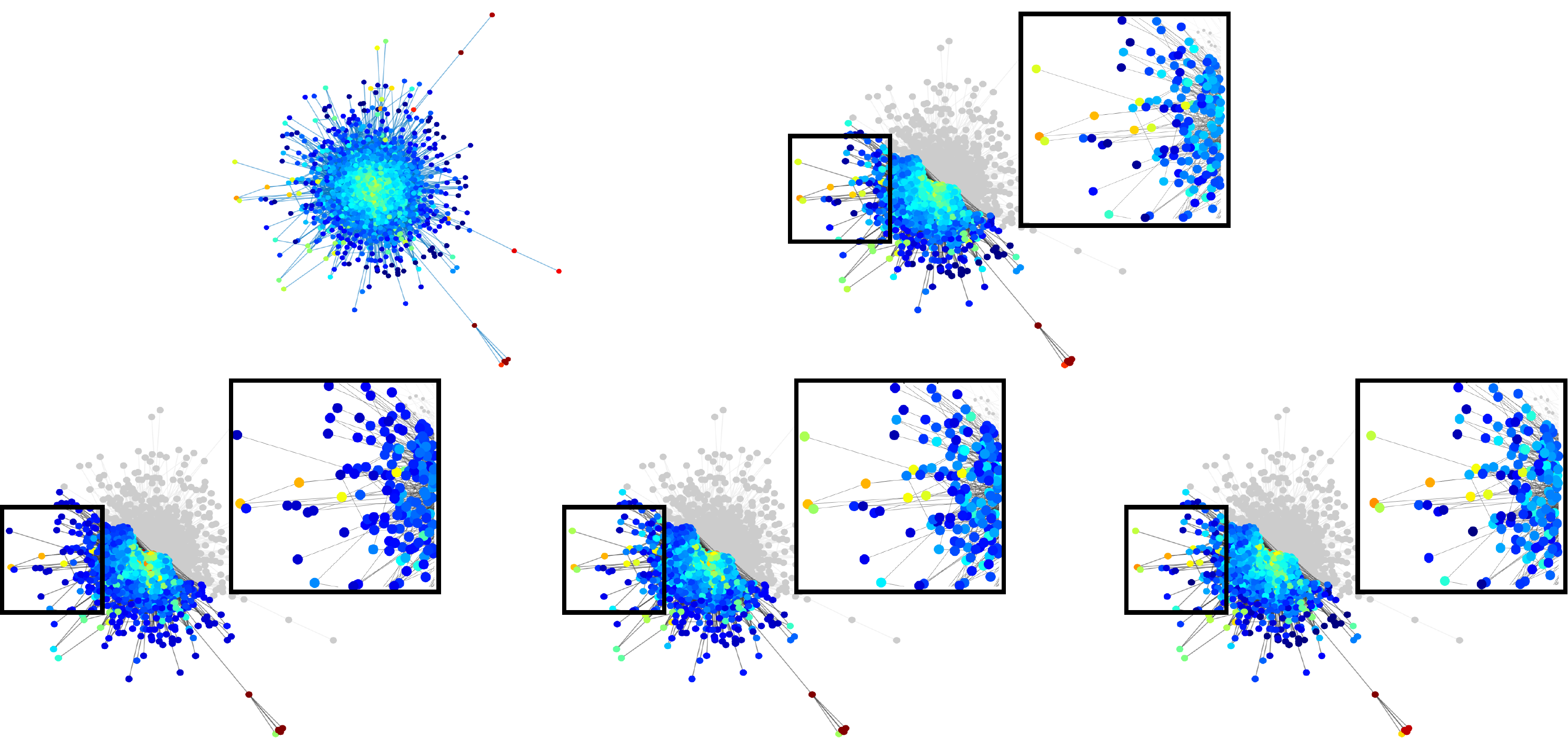}
\put(10,47){\footnotesize Original graph}
\put(50,47){\footnotesize Sub graph}
\put(8,22){\footnotesize 1\%}
\put(41,22){\footnotesize 10\%}
\put(78,22){\footnotesize 75\%}
\end{overpic}
\hfill
\begin{overpic}
[trim=-1cm -1.1cm 0 0cm,clip,width=0.4\linewidth,grid=false]{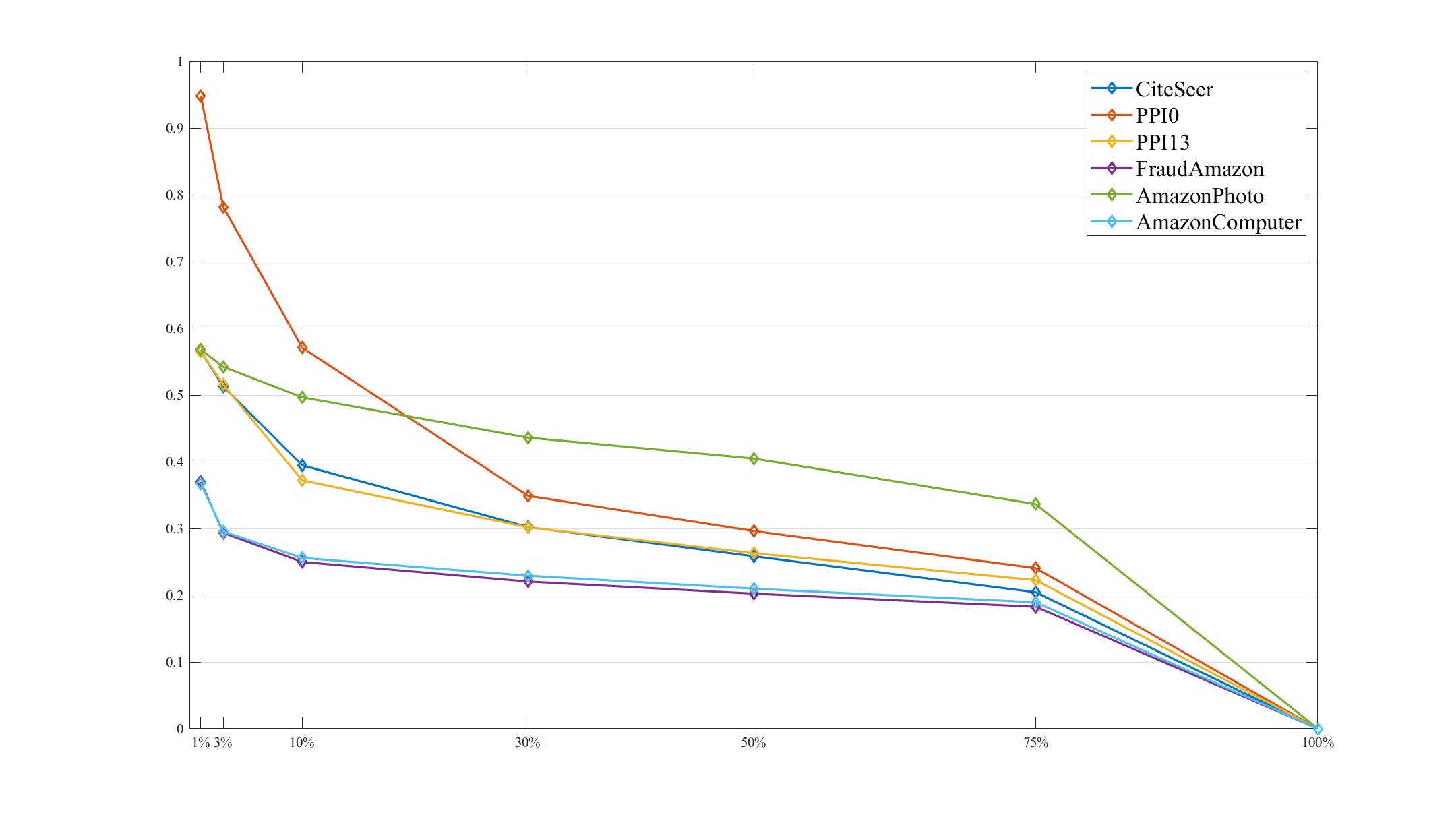}
\put(25,3){\footnotesize Percentage of eigenfunctions}
\put(2,35){\rotatebox[origin=c]{90}{\footnotesize RMSE}}
\end{overpic}
\caption{\label{fig:func_transfer} RMSE obtained by transferring positional encodings (PE) using the spectral map with an increasing amount of eigenfunctions. On the left, we show a qualitative example of signal transfer on PPI0. The first row shows the full graph and the partial graph, with the PE plotted on top. The bottom row shows the results of signal transfer with different percentage of eigenfunctions. On the right, we plot the RMSE at increasing percentages of eigenfunctions. }
\end{figure*}
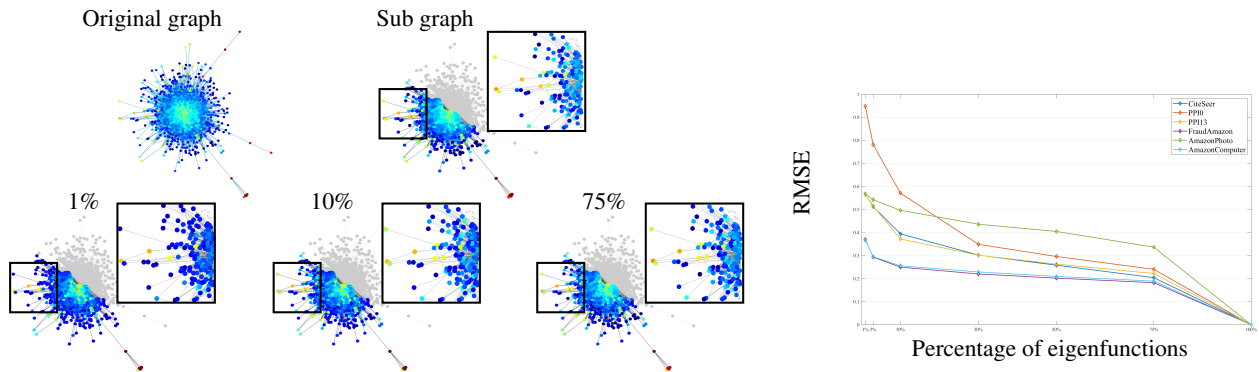

%% file: sections/conclusion.tex
\section{Conclusions}
\label{sec:conclusion}

The spectral representation of maps for encoding graph and subgraph maps lends itself well to several applications, and we anticipate that it will be a useful addition to the graph learning toolset.

Further, while in this paper we showed extensive evidence that the spectral map representation bears a special structure depending on the type of partiality, currently we have not taken full advantage of this structure. When the task at hand requires seeking for the subgraph alignment, i.e. whenever the map is unknown, it may be possible to design stronger regularizers to induce sparsity in the matrix representation of the map. This is quite different from the better known setting of 3D surfaces, where this sparse structure is typically just diagonal or slanted-diagonal. 

In the light of the increasing interest of the graph learning community toward spectral techniques, adopting a spectral representation for maps between graphs is a natural next step; it is simple to adopt, easy to manipulate, and memory-efficient, and has the potential to become a fundamental ingredient in spectral graph learning pipelines in the near future.


%% file: sections/Appendix/interpretation.tex
\section{Interpretation of the spectral map matrix}\label{sec:interp}

\subsection{\label{sec:fmpaSurf}Functional maps on surfaces}
Consider two smooth manifolds $\M$ and $\N$, and let $T:\N\to\M$ be a point-to-point map between them. Given a scalar function $f:\M\to\mathbb{R}$, the map $T$ induces a functional mapping via the composition $g = f \circ T$, which can be seen as a linear map $T_F:f\mapsto g$ from the space of functions on $\M$ to the space of functions on $\N$. 
As a linear map, the functional $T_F$ admits a matrix representation after choosing a basis for the two function spaces. 

To this end, consider a discretization of $\M$ and $\N$, with vertices $\Vg$ and $\Vs$ respectively, and the corresponding discretized version of their Laplace-Beltrami operators (LBOs) {(the counterpart of the graph Laplacian on smooth manifolds)}. The first $k$ eigenfunctions of the two LBOs can be concatenated side by side as columns to form the matrices $\Phi \in \mathbb{R}^{\vert \Vg \vert \times k}$ and $\Psi \in \mathbb{R}^{\vert \Vs \vert \times k}$.
%
%
Further, assume the pointwise map $T$ is available and encoded in a binary matrix $S$, such that $S(y,x) = 1$ if $y \in \Vs$ corresponds to $x \in \Vg$, and $0$ otherwise. By choosing $\Phi$ and $\Psi$ as bases, the functional map $T_F$ can be encoded in a small $k \times k$ matrix $C$ via the change of basis formula:
\begin{align}
\label{eq:fmap}
     C = \Psi^{\dagger} S \Phi\,,
\end{align}
where $\dagger$ is the Moore-Penrose pseudoinverse. The size of $C$ does {\em not} depend on the number of points in $\M$ and $\N$, but only on the number $k$ of basis functions. In other words, $C$ represents the linear transformation that maps the coefficients of any given function $f:\M\to\mathbb{R}$ expressed in the eigenbasis $\Phi$, to coefficients of a corresponding function $g:\N\to\mathbb{R}$ expressed in the eigenbasis $\Psi$.

When the pointwise similarity $S$ is unknown, one can directly compute the matrix $C$ as the solution of a regularized least-squares problem with $k^2$ unknowns, given some input features on the two surfaces (e.g., landmark matches or local descriptors). 
%
For further details we refer to \cite{ovsjanikov2012functional,ovsjanikov2017computing}.

\subsection{\label{sec:smooth}Comparison with smooth surfaces}
In the case of smooth surfaces, it has been shown~\citep{rodola2017partial} that the sparsity pattern of matrix $C$ can be well approximated by a simple formula. Given a surface $\M$ and an isometric part $\N$, the matrix $C$ is approximately diagonal, with diagonal angle $\alpha$ proportional to the ratio of surface areas:
\begin{align}
    \alpha \sim \frac{\mathrm{Area}(\N)}{\mathrm{Area}(\M)}\,.
\end{align}
As a a special case, full-to-full isometric shape matching yields a diagonal matrix $C$, since $\mathrm{Area}(\N)=\mathrm{Area}(\M)$. This result comes directly from an application of Weyl's asymptotic law for Laplacian eigenvalues of smooth manifolds~\citep{Weyl1911}, which relates the eigenvalue growth to the surface area of the manifold via the relation:
\begin{align}\label{eq:weyl}
    \lambda_\ell \sim \frac{(2\pi)^2}{\mathrm{Area}(\M)^{2/d}}\ell^{2/d}\,, \quad\quad\quad \ell\to\infty
\end{align}
where $d$ is the dimension of the manifold ($d=2$ for surfaces).
We refer to~\citealp[Eq. 9]{rodola2017partial} for additional details pertaining surfaces.

However, Weyl's law (Equation~\ref{eq:weyl}) does {\em not} hold for graphs, since there is not a well-defined notion of ``area'' of a graph. In fact, when we work with graphs and subgraphs, we observe that matrix $C$ does not necessarily follow a diagonal pattern. More general sparse structures are observed in the coefficients of $C$, but an explanation rooted in differential geometry is not readily available.

\begin{figure}[t!]
\centering
\begin{overpic}
[trim=0cm -0.2cm 0cm -2cm,clip,width=0.999\linewidth]{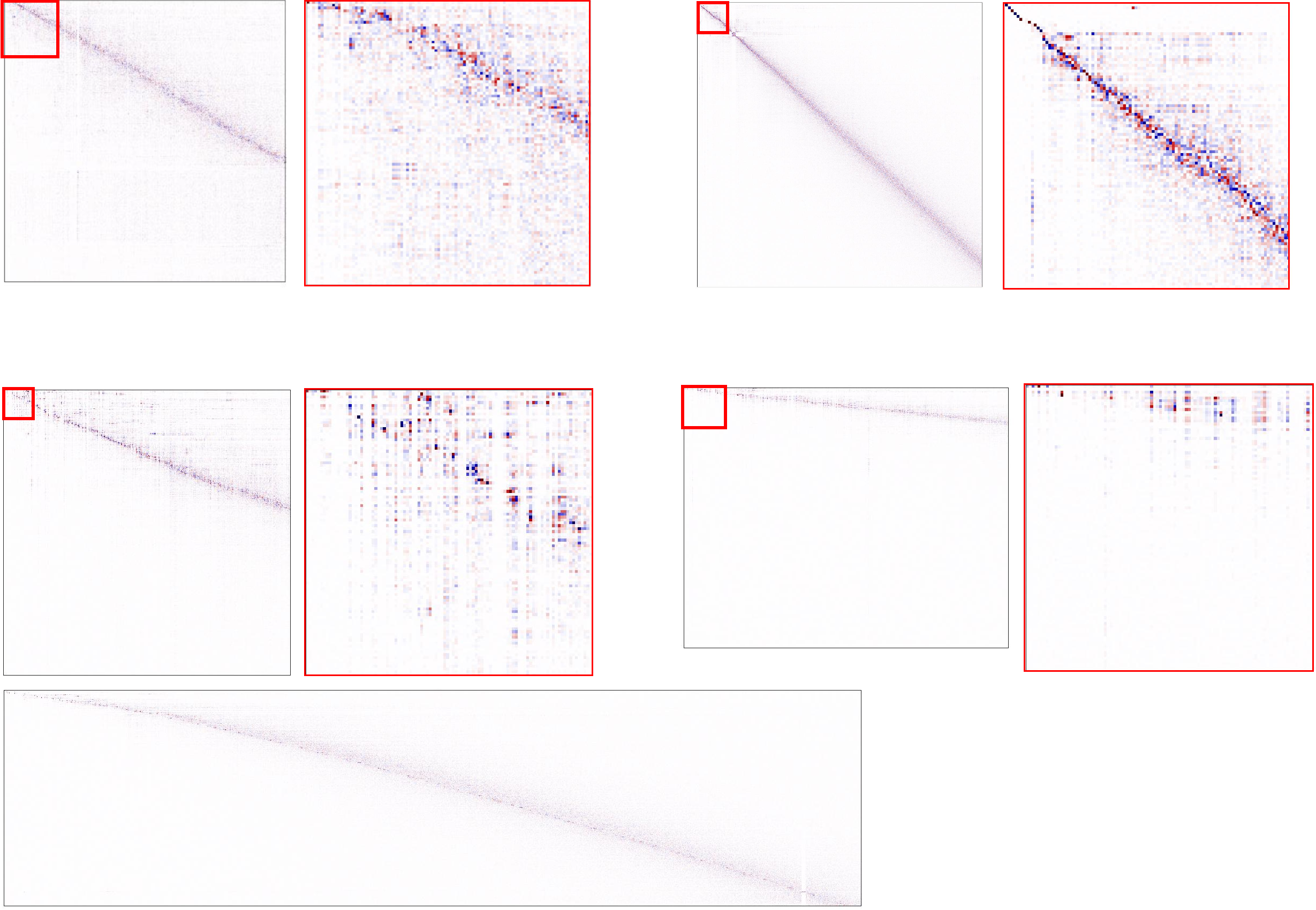}
\put(20,73){\tiny CORA}
\put(8,71){\tiny full graph: $2485$ nodes; subgraph: $1491$ nodes}
\put(-1.5,69.5){\tiny $1$}
\put(-3,66){\tiny $100$}
\put(-3,48.5){\tiny $500$}
\put(74,73){\tiny PPI0}
\put(61,71){\tiny full graph: $1546$ nodes; subgraph: $1428$ nodes}
\put(51.5,69.5){\tiny $1$}
\put(49.5,67.2){\tiny $100$}
\put(50.5,48.5){\tiny $1k$}
\put(18,44){\tiny Amazon Photo}
\put(9,42){\tiny full graph: $7487$ nodes; subgraph: $1609$ nodes}
\put(-1.5,40){\tiny $1$}
\put(-3,37.7){\tiny $100$}
\put(-2,19){\tiny $1k$}
\put(-1.5,16.5){\tiny $1$}
\put(-3,1.1){\tiny $940$}
\put(0.5,-0.5){\tiny $1$}
\put(63,-0.5){\tiny $3.7k$}
\put(70,44){\tiny Amazon Computer}
\put(62,42){\tiny full graph: $13381$ nodes; subgraph: $800$ nodes}
\put(50.5,40){\tiny $1$}
\put(48.5,37.2){\tiny $100$}
\put(48.5,21){\tiny $800$}
\end{overpic}
\caption{\label{fig:amazon}spectral maps computed over abstract graphs from 4 different datasets (CORA~\citep{cora}, PPI0~\citep{PPI}, Amazon Photo~\citep{AmazonCoBuy} and Amazon Computer~\citep{AmazonCoBuy}), showing a clear pattern in all cases. For each dataset, we compute the spectral map matrix $C$ between the complete graph and a subgraph; the subgraph is obtained according to a semantic criterion depending on the dataset, e.g., for Amazon Photo, by considering the subgraph of nodes belonging to the same product category. For each spectral map matrix $C$, we also show a zoom-in (framed in red). All the matrices are sparse, and have a clean structure that in some cases approximates a slanted diagonal. The wide matrix on the bottom is computed on Amazon Photo (using a different subgraph than the one used in the example above it), and shows that the sparse behavior is maintained throughout the entire spectrum.}
\end{figure}

In Figure~\ref{fig:amazon}, we report additional examples with large abstract graphs undergoing partiality transformations, showing that clear patterns appear rather consistently across different datasets.

Based on these observations, we believe there is an intriguing theoretical gap between what has been observed in the case of smooth manifolds, and what we report for graphs in this paper. In the former case, a geometric explanation has been proposed in the literature. In the latter case, empirical evidence yields similar results, yet it seems to be a purely algebraic phenomenon that remains to be addressed.

\subsection{\label{sec:nume_eig}Number of eigenvectors}
Given two graphs $G_1$ and $G_2$ with $m$ and $n$ nodes respectively, the node-to-node map $S$ has size $n \times m$, thus scaling quadratically with the number of nodes. 

By contrast, matrix $C$ as defined in Equation~\ref{eq:Cg} has dimensions that only depend on the number of Laplacian eigenvectors encoded in the matrices $\Phi_1,\Phi_2$. If one chooses the first $k_1 \ll m$ Laplacian eigenvectors for $G_1$ and the first $k_2 \ll n$ Laplacian eigenvectors for $G_2$, the size of $C$ is $k_2\times k_1$. Observe that $C$ is rectangular in general, but can be made square by choosing $k_1=k_2$ if so desired.

\begin{figure}[t]
\centering
\begin{overpic}
[trim=0cm 0cm 0cm 0cm,,clip,width=0.6\linewidth, grid=false]{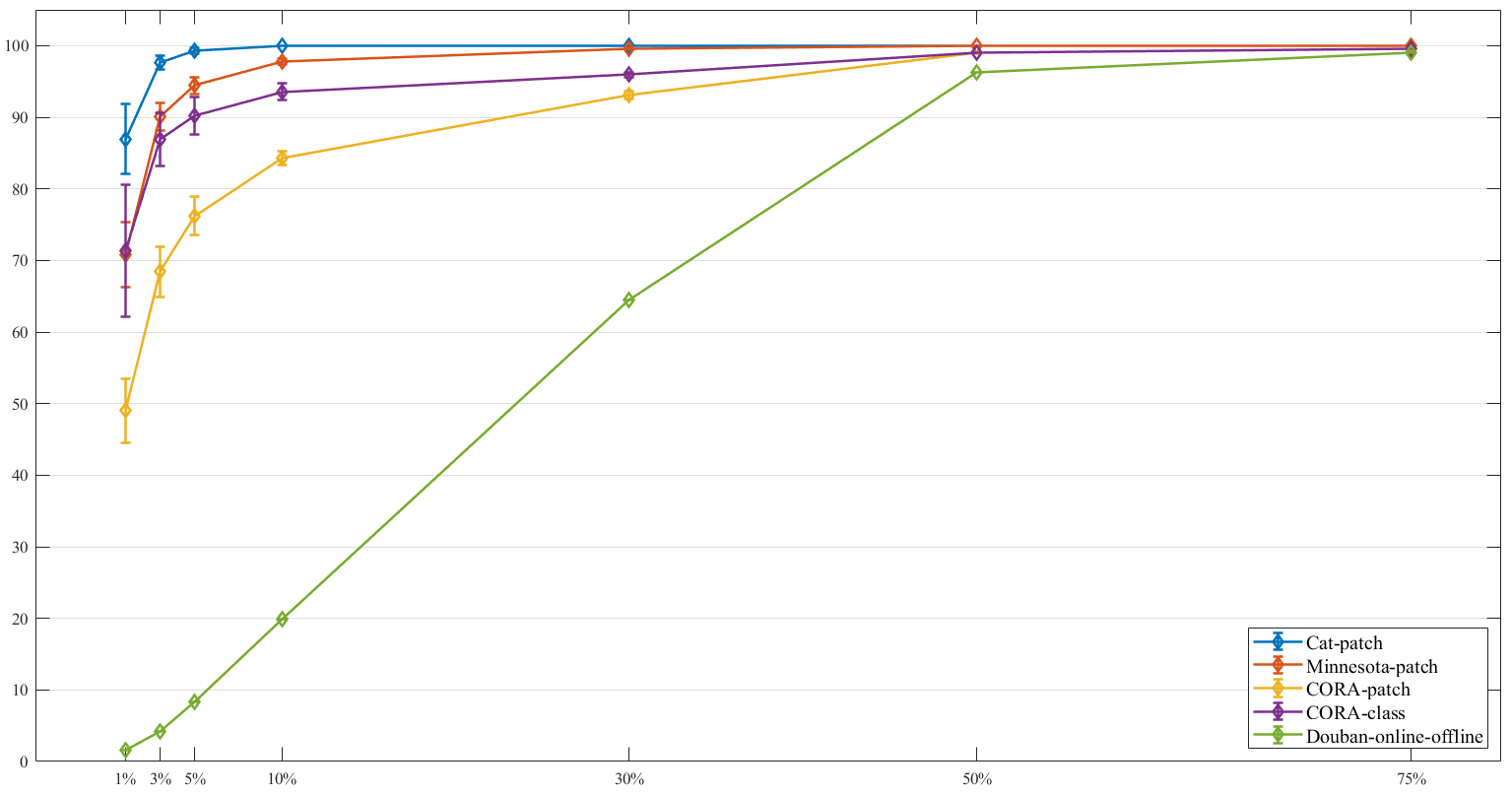}
\put(38,-2){\scriptsize Eigenvectors percentage}
\put(-3,26){\rotatebox[origin=c]{90}{\scriptsize MAP(\%)}}
\end{overpic}
\caption{\label{fig:n2n_eigs} MAP(\%) of the correspondence on different datasets at increasing number of eigenvectors (expressed as percentages, growing from $1\%$ to $75\%$). The correspondences are obtained from ground-truth spectral maps.}
\end{figure}


The experiments in Figure~\ref{fig:func_transfer} and ~\ref{fig:n2n_eigs} show that as the number of eigenvectors increases, the performance also increases. 
The Mean Average Precision (MAP) is defined as $\frac{1}{n} \sum_{i=1}^{n} \frac{1}{ra_i}$ where $ra_i$ is the rank (position) of positive matching node in the sequence of sorted candidates.
In particular, Figure~\ref{fig:n2n_eigs} demonstrates that, in most of the cases, a low percentage of eigenvectors (about 5\%) suffices to retrieve a good node-to-node correspondence; while at 50\% of the eigenvectors on all graphs the error is above 90\%.
As a general guideline, in this paper we typically use $k=20\sim 50$ for a graph with $1000$ nodes, leading to an especially compact representation $C$. 

\subsection{Choice of Laplace operator}


\begin{wrapfigure}[10]{r}{0.45\linewidth}
\vspace{-1.3cm}
\begin{center}
\begin{overpic}
[trim=0cm 0cm 0cm 0cm,clip,width=0.95\linewidth, grid=false ]{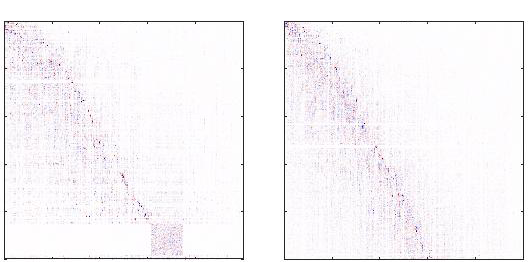}
\put(13,47){\tiny $L = D-A$}
\put(52,47){\tiny $\mathcal{L} = I-D^{-1/2}AD^{-1/2}$}
\end{overpic}
\end{center}
\caption{\label{fig:Lapl_choice} Spectral maps computed with two different Laplacians between the CORA graph and its subgraph.}
\end{wrapfigure}
A spectral map can be computed from the eigenbasis of any linear operator. In this paper we  use the symmetrically normalized graph Laplacian $\mathcal{L} = I-D^{\frac{1}{2}} A D^{\frac{1}{2}}$. A valid alternative is the standard Laplacian $L = D-A$, which shows similar behavior to the normalized counterpart. At a practical level, we observed that the Laplacian $L$ suffers from more problems of high multiplicity at lower frequencies, see Figure \ref{fig:Lapl_choice}.  

In the special case where the graph is constructed on top of a  point cloud sampled from a (possibly high-dimensional) manifold $\M$, it has been shown that the eigenvectors of the normalized graph Laplacian converge to the eigenfunctions of the Laplace-Beltrami operator on $\M$~\citep{BelkinN06}. However, as discussed in Appendix~\ref{sec:smooth}, our case is more general. We consider generic abstract graphs without an explicit underlying manifold, i.e. we do not construct our graphs from input point clouds. Further, in~\cite{BelkinN06} it is assumed that $\M$ is a compact infinitely differentiable Riemannian submanifold of $\mathbb{R}^d$ {\em without} boundary, meaning that partiality transformations, which are the main focus of this paper, are not considered.

%% file: sections/Appendix/exp_details.tex
\section{Dataset and implementation details}
\label{sec:exp_details}
In this section we report additional details about the experimental setup used in the main manuscript. 

\subsection{Datasets}
In Table~\ref{tab:datasets} we sum up the main statistics across all the datasets and benchmarks used in our experiments. In addition to number of nodes, number of edges, graph diameter and average node degree, in the table we also report the application domain of each dataset, the task where they are used, the type and number of node-wise features (where used).
Since PPI and QM9 are collections of graphs, we used only a subset. In particular, from the PPI dataset we used the first and fourteenth graphs (specified with 0 and 13 in the experiments). 
The Cat graph is derived from the corresponding mesh of the SHREC'16 Partial Deformable Shapes benchmark~\citep{cosmo2016shrec}. 

\input{sections/Tabels/TAB_datasets}

\input{sections/Figs/graph_rewiring}

\subsection{\label{app:rob_rewire}Robustness to rewiring}

In this Section, we formally define the connectivity changes and spectral map robustness used in Section \ref{sec:RobRewire}.
Given two graphs $G=(V, E)$ and $G'=(V', E')$, we measure the amount of change from $G$ to $G'$ as the (minimum) number of edits needed to transform $E$ to $E'$, divided by $|E|$: $\frac{(|E-E'| + |E'-E|)}{|E|}$. 
In our experiments, we consider small changes in the graph connectivity as a perturbation of  3\% of the edges. The rewiring operation we applied to the graphs consists of the deletion or addition of the same number of edges.

{We define the difference between the spectral map $C$ and $C'$ as $\| C- C' \|_F^2$. Note that there is ambiguity in the sign of the eigenfunctions of $C'$; to factor it out from the error computation, we use the sign that minimizes the error.}

{In Figure \ref{fig:graphsRewiring} we show the spectral maps generated from the experiment in Figure \ref{fig:robustRewiring}. Figure \ref{fig:CgraphsRewiring} shows the spectral map between the full and partial graphs from 0\% to 30\% of rewiring; Figure \ref{fig:deltaGraphsRewiring} shows the variation in the functional representation between the non-rewired case and the different percentages of rewiring.}

%% file: sections/Tabels/TAB_datasets.tex
\begin{table}[h]
\centering
\caption{Summary of statistics about the datasets used in our experiments.}
\tiny
\setlength{\tabcolsep}{2pt}
\begin{tabular}{l|cccC{0.9cm}cc cC{0.9cm}}
Dataset   & Nodes & Edges & Diameter & Average degree & Domain & Task & Features & Number of features\\ \hline
QM9  \citep{QM9}    &  29 & 47  & 6 & 3.24    &    Chemistry    &  Graph regression&-&-   \\
Karate \citep{karate}&34 & 78 & 5 & 4.59 &  Social networks   &  Node classification  & - & -  \\
PPI 0 \citep{PPI}      &    1546 & 17699  & 8 & 21.90  &   Chemistry     &   Graph regression & Gene attributes & 50 \\
Citeseer  \citep{CiteSeer}    &   2120 & 3731  & 28 & 3.50  &      Citation networks    & Node classification  & Bag-of-Words & 3703 \\
Cora   \citep{cora}   & 2485 & 5069 & 19 & 4.08 &    Citation networks    &  Node classification & - & -  \\
Minnesota &       2635  &   3298    & 98 & 2.5  &  Roadmap & - &- & 1,433   \\
PPI 13 \citep{PPI}  &   3480 & 56857  & 8 & 31.68   &   Chemistry     &   Graph regression & Gene attributes & 50 \\
Douban \citep{Douban}     &  3906 & 8164  & 13 & 4.18  &    Social networks  & Network alignment &  - & - \\
Amazon Photo \citep{AmazonCoBuy}  & 7487 & 119044  & 11 & 31.80   &        Co-purchase          &    Node classification  &    Bag-of-Words    &   745  \\
Cat \citep{shrec2016}       &   10000    &   19940    &    86 & 5.99    & Geometry processing  & Shape matching  & - & - \\
FraudAmazon  \citep{AmazonFraud}    &     11944 & 4417576  & 4 & 739.71    &   Product reviews & Fraud detection  & Bag-of-Words & 25\\
Amazon Computer \citep{AmazonCoBuy}  & 13381 & 245778  & 10 & 36.74   &       Co-purchase           &  Node classification &   Bag-of-Words     &   767  \\
Coauthor-cs \citep{}  & 18333 & 163,788  & 24 & 8.93   &    Citation networks       &  Node classification &    Bag-of-Words    &   6,805  \\
Pubmed \citep{}  & 19717 & 88,648  & 18 & 4.5     &      Citation networks     &  Node classification &   Bag-of-Words      &   500  \\
\end{tabular}
\label{tab:datasets}
\end{table}

%% file: sections/Figs/graph_rewiring.tex
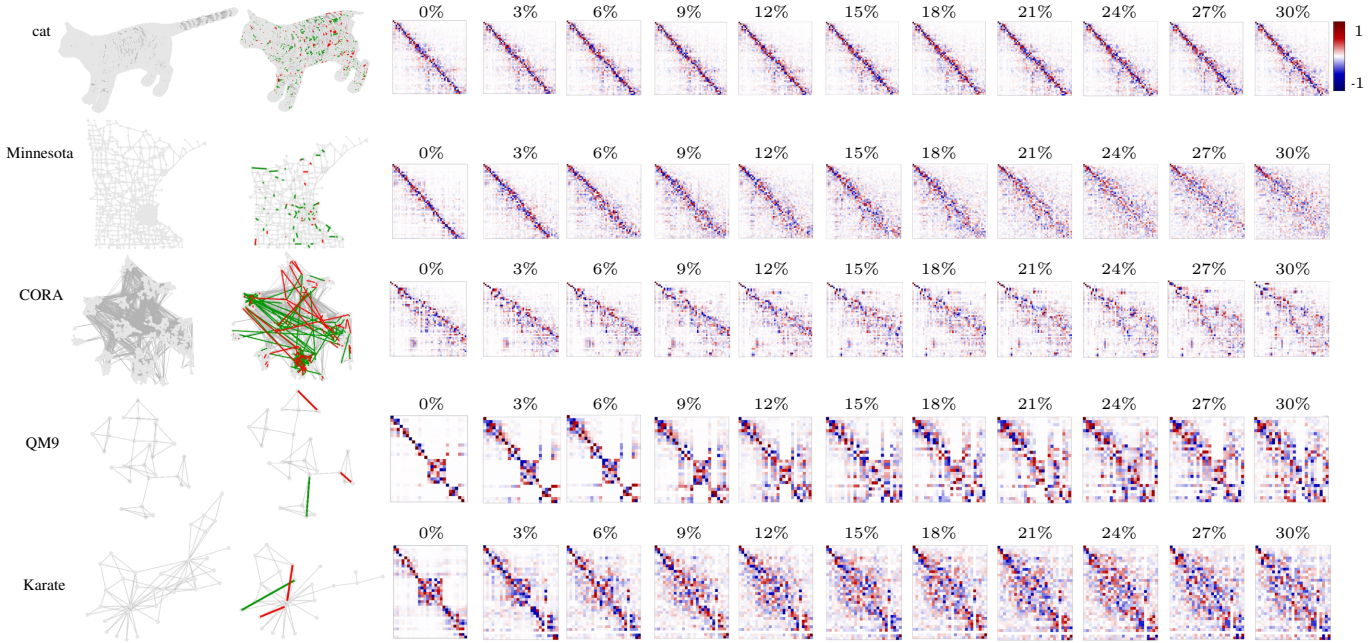
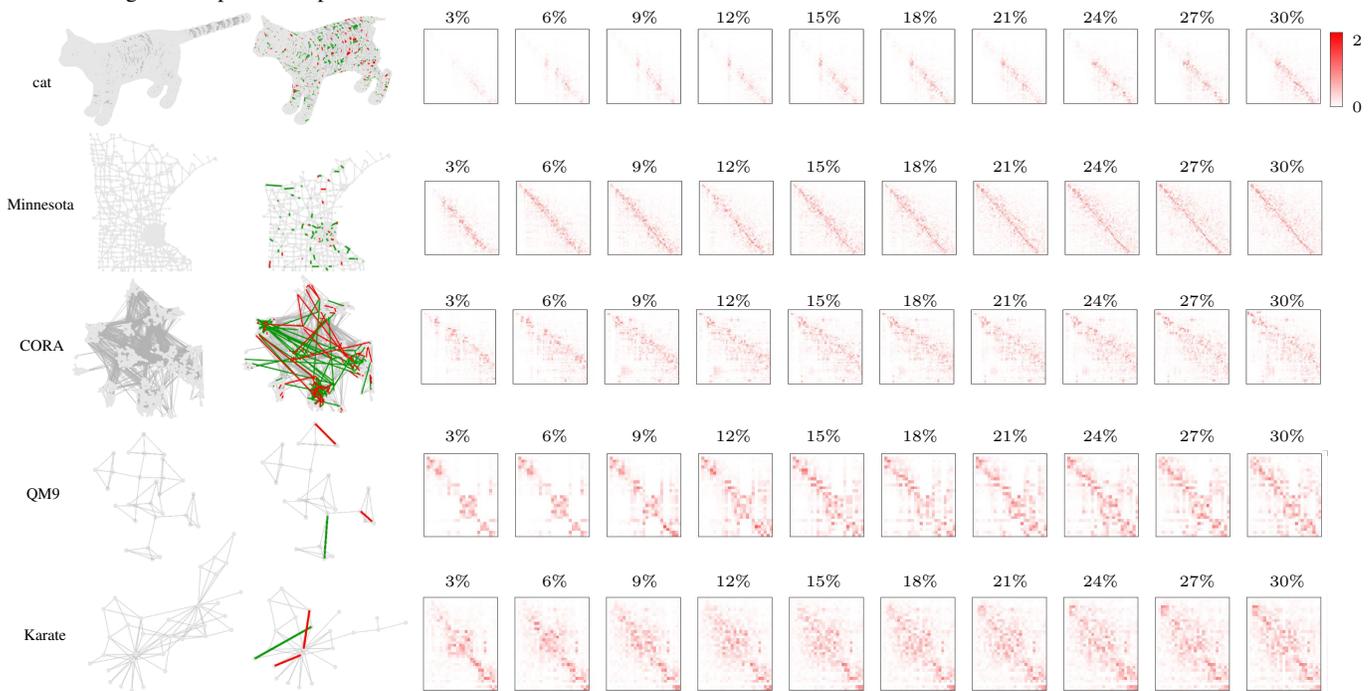
\begin{figure}[p]
\vspace{0.2cm}
\centering
\begin{subfigure}[b]{\textwidth}
\begin{overpic}
[trim=0cm 0cm 0cm 0cm,clip,width=0.999\linewidth, grid=false]{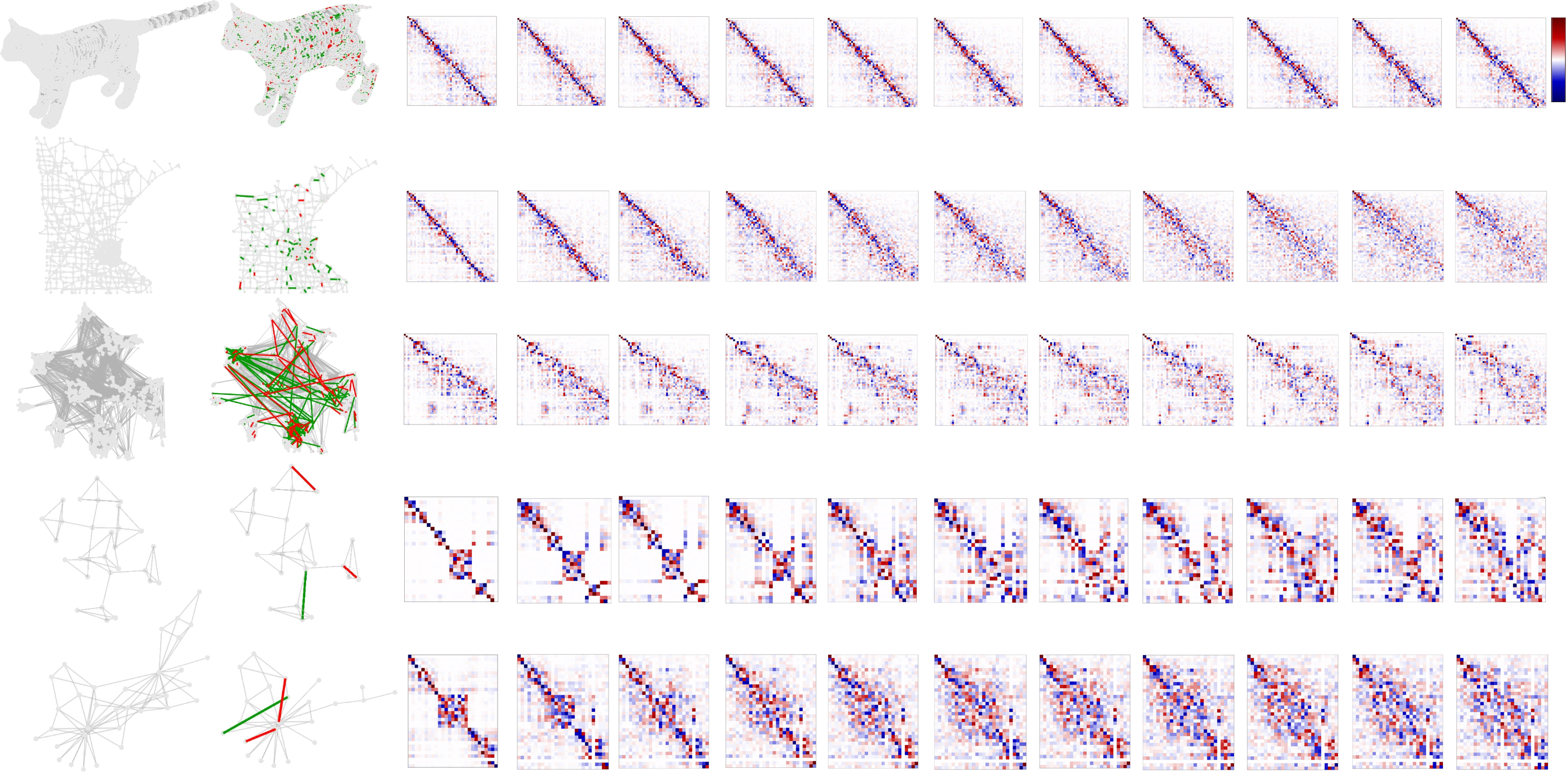}
\put(-2,47){\tiny cat}
\put(-4,37.5){\tiny Minnesota}
\put(-3,26.5){\tiny CORA}
\put(-2.5,15){\tiny QM9}
\put(-2.7,4){\tiny Karate}
\put(100.6,47){\tiny $1$}
\put(100.4,43){\tiny $\scalebox{0.7}[1.0]{-}1$}
\put(28,48.5){\tiny $0\%$}
\put(35.3,48.5){\tiny $3\%$}
\put(41.5,48.5){\tiny $6\%$}
\put(48,48.5){\tiny $9\%$}
\put(54,48.5){\tiny $12\%$}
\put(61,48.5){\tiny $15\%$}
\put(67,48.5){\tiny $18\%$}
\put(74.5,48.5){\tiny $21\%$}
\put(81,48.5){\tiny $24\%$}
\put(88,48.5){\tiny $27\%$}
\put(94.5,48.5){\tiny $30\%$}
\put(28,37.5){\tiny $0\%$}
\put(35.3,37.5){\tiny $3\%$}
\put(41.5,37.5){\tiny $6\%$}
\put(48,37.5){\tiny $9\%$}
\put(54,37.5){\tiny $12\%$}
\put(61,37.5){\tiny $15\%$}
\put(67,37.5){\tiny $18\%$}
\put(74.5,37.5){\tiny $21\%$}
\put(81,37.5){\tiny $24\%$}
\put(88,37.5){\tiny $27\%$}
\put(94.5,37.5){\tiny $30\%$}
\put(28,28.5){\tiny $0\%$}
\put(35.3,28.5){\tiny $3\%$}
\put(41.5,28.5){\tiny $6\%$}
\put(48,28.5){\tiny $9\%$}
\put(54,28.5){\tiny $12\%$}
\put(61,28.5){\tiny $15\%$}
\put(67,28.5){\tiny $18\%$}
\put(74.5,28.5){\tiny $21\%$}
\put(81,28.5){\tiny $24\%$}
\put(88,28.5){\tiny $27\%$}
\put(94.5,28.5){\tiny $30\%$}
\put(28,18){\tiny $0\%$}
\put(35.3,18){\tiny $3\%$}
\put(41.5,18){\tiny $6\%$}
\put(48,18){\tiny $9\%$}
\put(54,18){\tiny $12\%$}
\put(61,18){\tiny $15\%$}
\put(67,18){\tiny $18\%$}
\put(74.5,18){\tiny $21\%$}
\put(81,18){\tiny $24\%$}
\put(88,18){\tiny $27\%$}
\put(94.5,18){\tiny $30\%$}
\put(28,8){\tiny $0\%$}
\put(35.3,8){\tiny $3\%$}
\put(41.5,8){\tiny $6\%$}
\put(48,8){\tiny $9\%$}
\put(54,8){\tiny $12\%$}
\put(61,8){\tiny $15\%$}
\put(67,8){\tiny $18\%$}
\put(74.5,8){\tiny $21\%$}
\put(81,8){\tiny $24\%$}
\put(88,8){\tiny $27\%$}
\put(94.5,8){\tiny $30\%$}
\end{overpic}
\caption{\label{fig:CgraphsRewiring} 
The plotted matrices represent the spectral map between the full and partial graphs from 0\% to 30\% of rewiring, showing the effect of rewiring on the spectral map structure. }
\end{subfigure}
\vfill
\begin{subfigure}[b]{\textwidth}
\vspace{0.2cm}
\centering
\begin{overpic}
[trim=0cm 0cm 0cm 0cm,clip,width=0.999\linewidth]{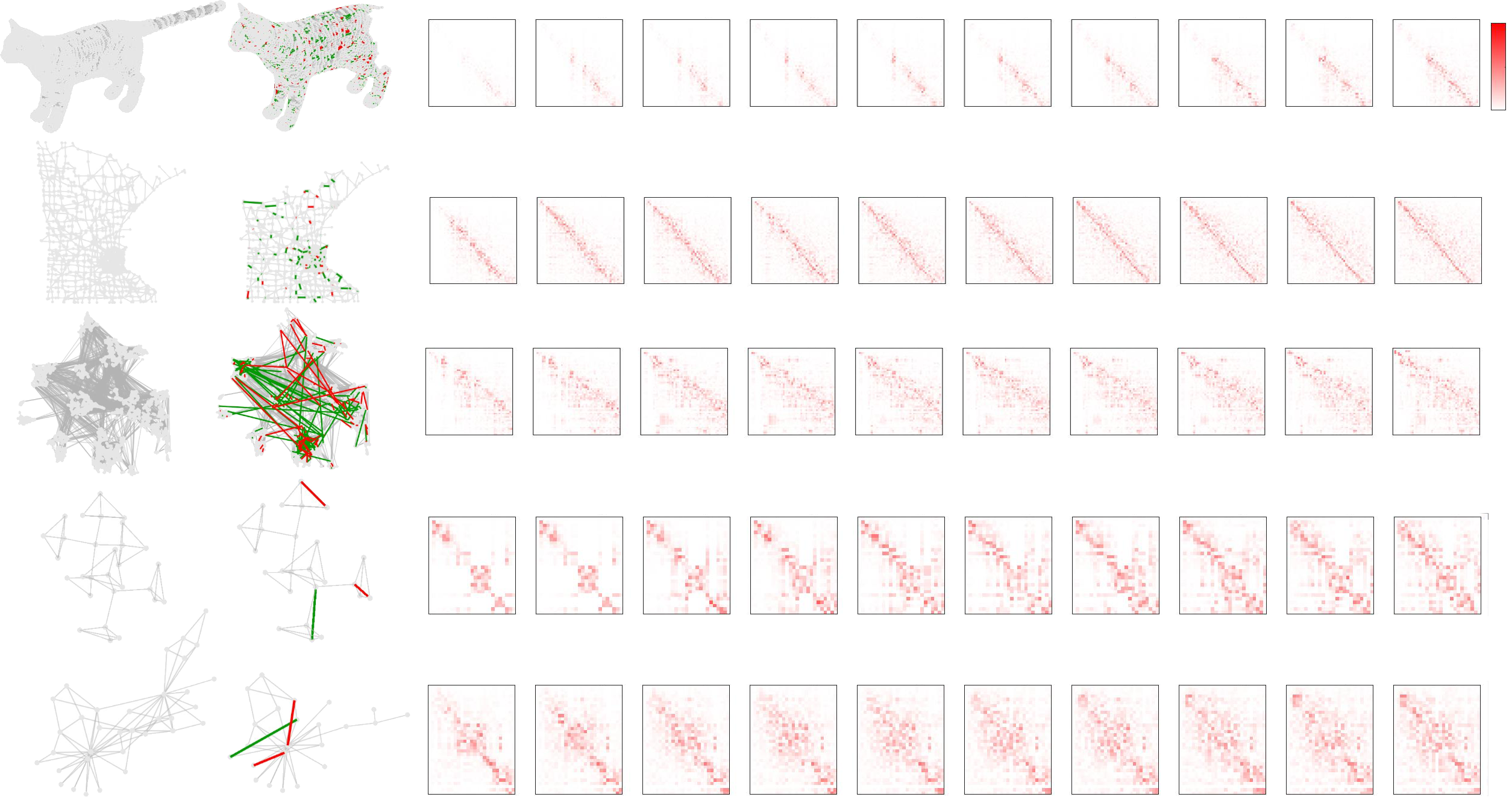}
\put(-2,47){\tiny cat}
\put(-4,37.5){\tiny Minnesota}
\put(-3,26.5){\tiny CORA}
\put(-2.5,15){\tiny QM9}
\put(-2.7,4){\tiny Karate}
\put(100.4,50.2){\tiny $2$}
\put(100.4,45){\tiny $0$}
\put(30,52){\tiny $3\%$}
\put(37.5,52){\tiny $6\%$}
\put(44.5,52){\tiny $9\%$}
\put(51,52){\tiny $12\%$}
\put(58,52){\tiny $15\%$}
\put(65.5,52){\tiny $18\%$}
\put(72.5,52){\tiny $21\%$}
\put(79.5,52){\tiny $24\%$}
\put(87,52){\tiny $27\%$}
\put(94,52){\tiny $30\%$}
\put(30,40.4){\tiny $3\%$}
\put(37.5,40.4){\tiny $6\%$}
\put(44.5,40.4){\tiny $9\%$}
\put(51,40.4){\tiny $12\%$}
\put(58,40.4){\tiny $15\%$}
\put(65.5,40.4){\tiny $18\%$}
\put(72.5,40.4){\tiny $21\%$}
\put(79.5,40.4){\tiny $24\%$}
\put(87,40.4){\tiny $27\%$}
\put(94,40.4){\tiny $30\%$}
\put(30,30){\tiny $3\%$}
\put(37.5,30){\tiny $6\%$}
\put(44.5,30){\tiny $9\%$}
\put(51,30){\tiny $12\%$}
\put(58,30){\tiny $15\%$}
\put(65.5,30){\tiny $18\%$}
\put(72.5,30){\tiny $21\%$}
\put(79.5,30){\tiny $24\%$}
\put(87,30){\tiny $27\%$}
\put(94,30){\tiny $30\%$}
\put(30,19.5){\tiny $3\%$}
\put(37.5,19.5){\tiny $6\%$}
\put(44.5,19.5){\tiny $9\%$}
\put(51,19.5){\tiny $12\%$}
\put(58,19.5){\tiny $15\%$}
\put(65.5,19.5){\tiny $18\%$}
\put(72.5,19.5){\tiny $21\%$}
\put(79.5,19.5){\tiny $24\%$}
\put(87,19.5){\tiny $27\%$}
\put(94,19.5){\tiny $30\%$}
\put(30,8.2){\tiny $3\%$}
\put(37.5,8.2){\tiny $6\%$}
\put(44.5,8.2){\tiny $9\%$}
\put(51,8.2){\tiny $12\%$}
\put(58,8.2){\tiny $15\%$}
\put(65.5,8.2){\tiny $18\%$}
\put(72.5,8.2){\tiny $21\%$}
\put(79.5,8.2){\tiny $24\%$}
\put(87,8.2){\tiny $27\%$}
\put(94,8.2){\tiny $30\%$}
\end{overpic}
\caption{\label{fig:deltaGraphsRewiring} The plotted matrices encode the element-wise error of the spectral map after the topological perturbations. Error is encoded as color, growing from white to red.}
\end{subfigure}
\caption{\label{fig:graphsRewiring} Robustness of the map to the simultaneous action of partiality and rewiring of the subgraph. The rewiring operations are increasingly \textbf{stronger}, with increments of $3\%$ of the total number of edges (starting from $3\%$ and reaching $30\%$). 
The second column shows one representative example (per dataset) of such topological modifications, depicting the added edges in green, and the removed edges in red.
The plotted matrices represent the spectral map after the topological perturbations, showing the effect of rewiring on the spectral map structure. }
\end{figure}

%% file: sections/Appendix/GZoom_eigs.tex
\section{\label{app:GZoomeigs} Hierarchical Graph Embedding: additional results}

\begin{figure}
    \centering
    \begin{subfigure}[b]{1\textwidth}
         \centering
         \includegraphics[trim={8cm 0 8cm 1cm},clip, width=\textwidth]{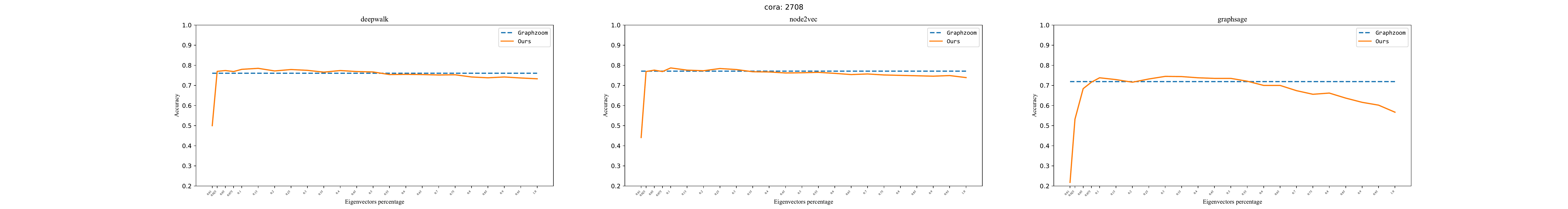}
         \caption{\label{fig:HGEeigs cora} Cora}         
     \end{subfigure}
    \begin{subfigure}[b]{1\textwidth}
         \centering
         \includegraphics[trim={8cm 0 8cm 1cm},clip, width=\textwidth]{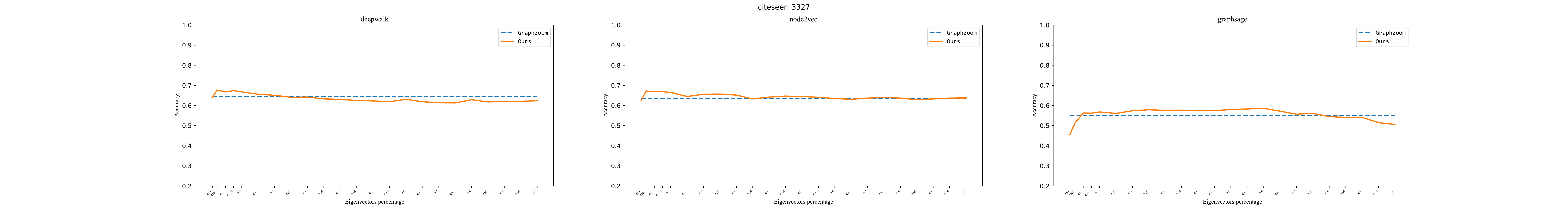}
         \caption{\label{fig:HGEeigs citeseer} Citeseer}         
     \end{subfigure}
     \begin{subfigure}[b]{1\textwidth}
         \centering
         \includegraphics[trim={8cm 0 8cm 1cm},clip, width=\textwidth]{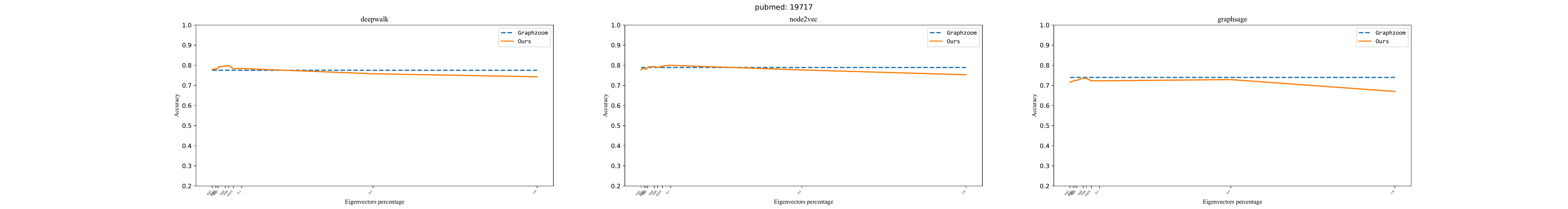}
         \caption{\label{fig:HGEeigs pubmed} Pubmed}         
     \end{subfigure}
    \caption{Node classification accuracy on the task of Hierarchical Graph Embedding at different percentages of eigenvectors}
    \label{fig:HGEeigs}
\end{figure}

In Figure \ref{fig:HGEeigs}, we report the accuracy performance for different percentages of eigenvectors in the experiment of Section \ref{sec:hiercEmb}. 
The performance of the spectral map rapidly increases at low percentages demonstrating the need for a few eigenvectors to obtain a good embedding lifting. When the percentages are higher than $50\%$ the accuracy decreases reaching the values of the node-to-node map at $100\%$. This phenomenon demonstrates that the spectral map can approximate the node-to-node map at  $100\%$ eigenvectors, but it is not the most convenient representation for the Hierarchical Embedding on graphs.

\section{\label{app:GKDadd} Geometric Knowledge Distillation: additional results}
\input{sections/Tabels/TAB_GKDtime.tex}

In Table \ref{tab:GKD_time} we show the mean epoch time registered during training. For each method and datset we report both the time in millisecond and the speed up compared to \textit{Ours ($5\%$)}. The spectral representation is able to reach a speed up of $200\%$ in some cases, demonstrating its convenience in terms of computation efficiency. 

%% file: sections/Tabels/TAB_GKDtime.tex
\begin{table}[t]

\caption{\label{tab:GKD_time} Computation time per epoch in the task of Knowledge Distillation. For each methods we report the mean time per epoch in milliseconds and the speed up with respect to Ours ($5\%$). For the spectral map, we report the performance with both $5\%$ and $10\%$ of eigenvectors.}
\centering
\tiny
\setlength{\tabcolsep}{3.5pt}
\begin{tabular}{l|cc|cc|cc|cc|cc|cc|cc}
              & \multicolumn{2}{c|}{Cora} & \multicolumn{2}{c|}{Citeseer} & \multicolumn{2}{c|}{Amazon-photo} & \multicolumn{2}{c|}{Amazon-computer} & \multicolumn{2}{c|}{Coauthor-cs} & \multicolumn{2}{c|}{Pubmed} & \multicolumn{2}{c}{Coauthor-physics} \\ \toprule
ORACLE& $2.47$ & $(54 \%)$& $2.68$ & $(54 \%)$& $3.75$ & $(47 \%)$& $6.27$ & $(40 \%)$& $5.22$ & $(51 \%)$& $2.75$ & $(50 \%)$& $10.84$ & $(48 \%)$\\
TEACHER& $2.47$ & $(54 \%)$& $2.68$ & $(54 \%)$& $3.75$ & $(47 \%)$& $6.27$ & $(40 \%)$& $5.22$ & $(51 \%)$& $2.75$ & $(50 \%)$& $10.84$ & $(48 \%)$\\
STUDENT& $2.81$ & $(47 \%)$& $2.95$ & $(49 \%)$& $3.18$ & $(55 \%)$& $4.32$ & $(59 \%)$& $5.24$ & $(51 \%)$& $2.84$ & $(48 \%)$& $9.24$ & $(56 \%)$\\
 \midrule
GKD-G& $7.28$ & $(-37 \%)$& $7.67$ & $(-32 \%)$& $25.79$ & $(-265 \%)$& $15.25$ & $(-45 \%)$& $18.80$ & $(-76 \%)$& $14.76$ & $(-169 \%)$& $39.66$ & $(-91 \%)$\\
GKD-R& $9.51$ & $(-79 \%)$& $9.86$ & $(-69 \%)$& $21.78$ & $(-208 \%)$& $16.85$ & $(-60 \%)$& $20.27$ & $(-90 \%)$& $16.46$ & $(-200 \%)$& $41.29$ & $(-99 \%)$\\

GKD-S& $6.27$ & $(-18 \%)$& $6.51$ & $(-12 \%)$& $17.31$ & $(-145 \%)$& $13.90$ & $(-32 \%)$& $17.45$ & $(-64 \%)$& $13.68$ & $(-149 \%)$& $38.35$ & $(-85 \%)$\\
PGKD& $7.90$ & $(-48 \%)$& $8.19$ & $(-40 \%)$& $17.53$ & $(-148 \%)$& $18.44$ & $(-75 \%)$& $20.36$ & $(-91 \%)$& $15.56$ & $(-183 \%)$& $45.02$ & $(-117 \%)$\\
 \midrule
Ours (5\%)& $5.32$ & $(0 \%)$& $5.83$ & $(0 \%)$& $7.07$ & $(0 \%)$& $10.53$ & $(0 \%)$& $10.66$ & $(0 \%)$& $5.50$ & $(0 \%)$& $20.77$ & $(0 \%)$\\
Ours (10\%)& $5.51$ & $(-4 \%)$& $6.18$ & $(-6 \%)$& $7.20$ & $(-2 \%)$& $10.63$ & $(-1 \%)$& $10.97$ & $(-3 \%)$& $5.83$ & $(-6 \%)$& $21.97$ & $(-6 \%)$\\
\end{tabular}
\end{table}